\documentclass[preprint,review,10pt]{elsarticle}
\usepackage[left=25mm,right=25mm,top=25mm,bottom=25mm,paper=a4paper]{geometry}

\usepackage{amssymb}
\usepackage{amsmath}

\usepackage{float}
\usepackage{graphicx}
\usepackage{array}
\usepackage{booktabs}
\usepackage{subfigure}
\usepackage{subcaption}
\usepackage{multirow}
\usepackage{tabularray}
\usepackage{arydshln} 
\usepackage{amsfonts}
\usepackage{amssymb}
\usepackage{amsmath}
\usepackage{adjustbox}
\usepackage{color, colortbl}
\usepackage{xcolor}
\usepackage{caption}

\usepackage{algpseudocode}
\usepackage{multirow}

\usepackage[ruled, linesnumbered, noend]{algorithm2e}
\usepackage{amsmath, amssymb}
\SetAlgoNlRelativeSize{-1}
\SetAlCapSkip{0.5em}

\journal{Pattern Recognition}

\begin{document}

\begin{frontmatter}

\title{ConMamba: Contrastive Vision Mamba for Plant \\Disease Detection}

\author[label1,label2]{Abdullah Al Mamun}
\author[label2]{Miaohua Zhang}
\author[label2]{David Ahmedt-Aristizabal}
\author[label2]{Zeeshan Hayder}
\author[label1]{Mohammad Awrangjeb}

\affiliation[label1]{organization={School of Information and Communication Technology},
            addressline={Griffith University},
            city={Nathan},
            postcode={4111},
            state={Queensland},
            country={ Australia.}}

\affiliation[label2]{organization={Imaging and Computer Vision Group},
            addressline={Data61, CSIRO},
            city={Black Mountain},
            postcode={2601},
            state={Canberra},
            country={Australia}}

\begin{abstract}

Plant Disease Detection (PDD) is a key aspect of precision agriculture. However, existing deep learning methods often rely on extensively annotated datasets, which are time-consuming and costly to generate.
Self-supervised Learning (SSL) offers a promising alternative by exploiting the abundance of unlabeled data. However, most existing SSL approaches suffer from high computational costs due to convolutional neural networks or transformer-based architectures. Additionally, they struggle to capture long-range dependencies in visual representation and rely on static loss functions that fail to align local and global features effectively. 
To address these challenges, we propose ConMamba, a novel SSL framework specially designed for PDD. ConMamba integrates the Vision Mamba Encoder (VME), which employs a bidirectional State Space Model (SSM) to capture long-range dependencies efficiently. Furthermore, we introduce a dual-level contrastive loss with dynamic weight adjustment to optimize local-global feature alignment. Experimental results on three benchmark datasets demonstrate that ConMamba significantly outperforms state-of-the-art methods across multiple evaluation metrics. This provides an efficient and robust solution for PDD. 
\end{abstract}



\end{frontmatter}


\section{Introduction} \label{sec:introduction}

\begin{figure}[H]
    \centering
    \includegraphics[width=0.15\textwidth]{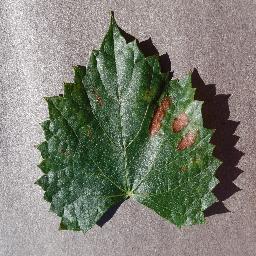}
    \includegraphics[width=0.15\textwidth]{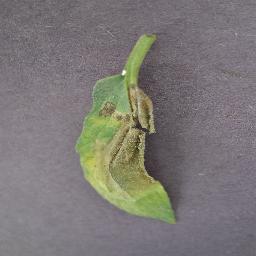}
    \includegraphics[width=0.15\textwidth]{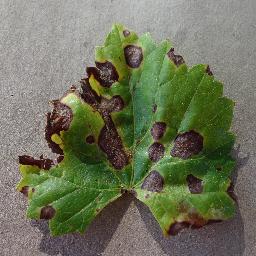}   
    \includegraphics[width=0.15\textwidth]{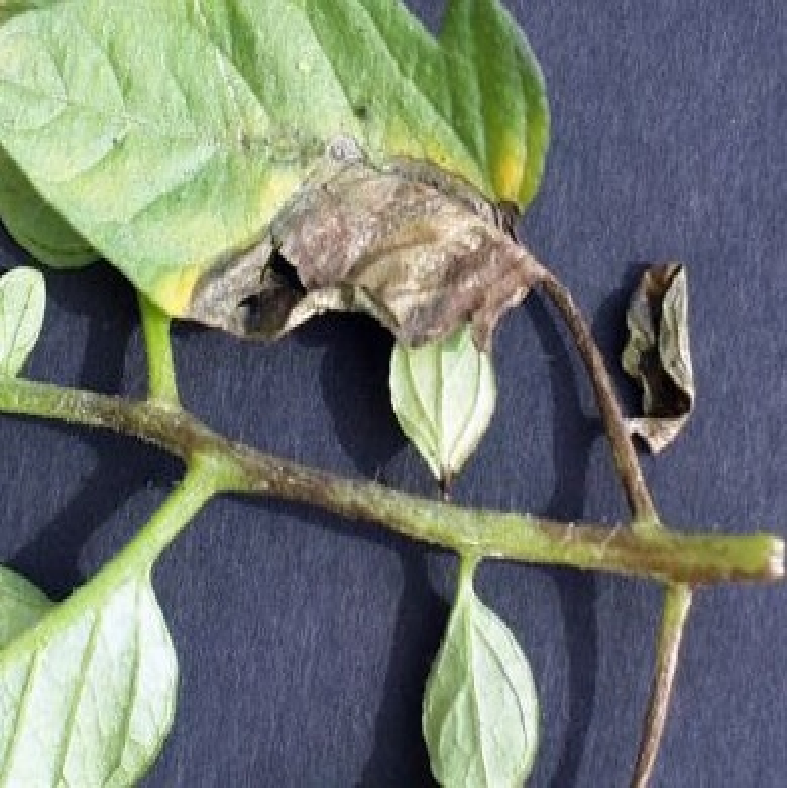}
    \includegraphics[width=0.15\textwidth]{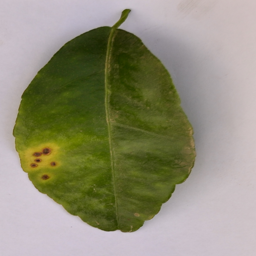}
    \includegraphics[width=0.15\textwidth]{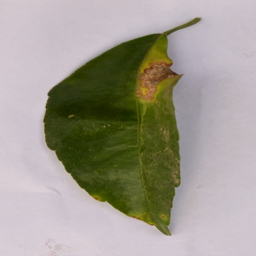}
    
    \caption{Sample images of affected plant leaves showing visible symptoms of the disease.}
    \vspace{-7pt}
    \label{fig:sample}
\end{figure}

Plant diseases pose a significant threat to crop growth and fruit production, leading to reductions in both yield and quality, resulting in major economic losses~\cite{jogekar2021review}. Studies indicate that plant diseases contribute to approximately 20\%-40\% of global crop yield losses \cite{gupta2019phytosanitary}. As illustrated in Figure \ref{fig:sample}, plant leaf diseases manifest in various forms. By closely examining leaf symptoms, accurate detection becomes possible, allowing for timely intervention and effective treatment \cite{dwivedi2021grape}.

Traditionally, plant disease diagnosis relied on human expertise, a time-consuming, labour-intensive process, and prone to errors due to the vast diversity of plant diseases~\cite{ngugi2021recent}. 
With the success of deep learning approaches across various domains, including agriculture \cite{ahmedt2024field,hossen2025transfer}, researchers have increasingly adopted these models to improve the accuracy and efficiency of plant leaf disease detection. Deep learning enables large-scale crop monitoring while reducing the costs associated with manual diagnosis \cite{ashwinkumar2022automated}. Nonetheless, challenges persist due to the complexity of plant disease identification and the limited availability of labeled training data.

Deep learning-based approaches for plant disease identification generally fall into three main categories: model enhancement~\cite{bansal2021disease}, Few-shot Learning (FSL) \cite{yang2022survey}, and Self-Supervised Learning (SSL) \cite{guldenring2021self}. Some studies also explore dataset augmentation using Generative Adversarial Networks (GANs) \cite{zhang2022mmdgan}, though this topic falls outside the scope of this discussion. Specifically, model enhancement primarily focuses on improving disease detection models by incorporating advanced architectures, such as Convolutional Neural Networks (CNNs) \cite{lecun1998gradient} and transformers \cite{wang2021pyramid}. While these approaches achieve high accuracy, they rely on supervised learning, which demands large amounts of labeled data. To address data scarcity, FSL strategies train models to identify similarities between samples and transfer this knowledge to new domains using only a small labeled dataset (support set). However, its effectiveness declines when the source domain differs considerably from the target dataset, limiting its broader applicability.

SSL has emerged as a powerful approach for Plant Disease Detection (PDD)~\cite{al2024plant}, addressing the challenges of limited labeled data by leveraging supervisory signals from raw, unlabeled datasets. Recent advancements in deep learning have significantly improved image-based disease identification, and the integration of SSL further enhances these capabilities by maximizing data utilization and improving model efficiency. By reducing reliance on annotated datasets, SSL offers a scalable and robust framework for modern PDD systems~\cite{Zhao2023CLA}.

Among SSL approaches, Contrastive Learning (CL) has emerged as the most effective and widely adopted method for vision-related tasks \cite{chen2020simple, hu2024comprehensive}. CL focuses on distinguishing positive pairs, such as augmented versions of the same image, from negative pairs, which are different images. This method has proven highly successful in capturing meaningful semantic representations, often achieving results comparable to or better than supervised learning on various downstream applications.

Traditional convolutional networks, such as ResNet50 \cite{he2016deep}, EfficientNetB0 \cite{Bunyang2023Self-supervised}, and VGG16 \cite{simonyan2014very}, are widely used for feature extraction in CL frameworks. However, they primarily capture local features, limiting their ability to represent global information. The transformer-based frameworks \cite{zhang2023information,yu2023mix} have better abilities to capture global information. However, the quadratic complexity of their inherent self-attention mechanism makes them computationally expensive.
 
These constraints hinder the efficient detection of complex, distributed disease symptoms in plants. Long-range dependencies are crucial for robust classification, as they enhance resilience to noise, occlusions, and environmental variations~\cite{zhang2022efficient}.
Additionally, static loss functions in CL restrict feature alignment across local and global scales, impacting adaptability and generalization to unseen disease scenarios~\cite{yang2022adaptive}. 

Recently, the Mamba framework \cite{gu2023mamba,gu2021efficiently}, which is a new deep learning architecture based on the SSM focuses on sequence modeling and offers several distinct advantages. By leveraging a state-space formulation, it efficiently captures long-range dependencies with linear computational complexity, which makes it more efficient than traditional self-attention mechanisms. This efficiency is especially beneficial when processing high-resolution plant disease images. Mamba's superior performance in capturing long-range dependencies is particularly important for PDD because disease symptoms can appear as subtle, scattered patterns across a leaf. By enhancing these long-range correlations, the model can integrate information
from various regions of the leaves/plants, thus producing more accurate disease predictions. For detailed information on Vision Mamba and its significance in PDD, readers are referred to Sections \ref{sec:vision mamba} and \ref{sec:Vm for PDD}, respectively.

Considering the advantages of the Mamba framework, we propose a Contrastive Mamba (ConMamba) framework to address the above-mentioned challenges for PDD. Specifically, a contrastive self-supervised learning framework based on Mamba is proposed by leveraging the Vision Mamba Encoder (VME) along with dual-view patch embedding techniques. This design enables the extraction of robust and rich visual representations by capturing both fine-grained local features as well as global context. Besides, the proposed framework incorporates dynamic feature
alignment across both local and global scales, effectively capturing intricate spatial patterns and long-range dependencies. This enhanced feature integration is important to
achieving robust and accurate PDD in complex real-world scenarios

The key contributions of this work include:
\begin{itemize}

\item A self-supervised Contrastive Vision Mamba (ConMamba) framework is developed specially tailored for plant disease detection. ConMamba integrates a bidirectional State Space Model (SSM) to efficiently capture long-range contextual relationships, enabling robust visual representation of subtle, localized symptoms and complex global disease patterns.

\item A dual-level contrastive learning strategy consisting of intra/inter-class contrastive learning is developed to both promote local embedding consistency across augmented image views and enforce global discriminability between embeddings of distinct disease classes. The dual approach ensures comprehensive representation learning, addressing the critical need for effective local-global feature alignment.
\item To adaptively balance the intra- and inter-contrastive losses during training, a dynamic weighting mechanism based on learnable uncertainty parameters is designed. This mechanism significantly enhances training robustness and generalization performance, resulting in highly discriminative embeddings suitable for accurate disease classification.

\end{itemize}

The remainder of the paper is organized as follows: 
Section \ref{sec:Related works} presents a literature review and motivation on PDD using self-supervised CL. 
Section \ref{sec:Preliminaries} explains the preliminaries of VM, including its advantages on plant diseases. 
Section \ref{sec:method} details the proposed ConMamba framework. 
Section \ref{sec:experiment} evaluates ConMamba on three benchmark datasets: PlantVillage, PlantDoc, and Citrus. 
Finally, Section \ref{sec:Conclusion} concludes the work and discusses directions for future work.


\section{Related Works and Motivations}\label{sec:Related works}
\subsection{Self-supervised Learning}

SSL has emerged as a transformative paradigm for PDD, addressing the critical challenge of limited annotated datasets. Unlike supervised methods that rely heavily on expensive and time-consuming manual labeling, SSL extracts supervisory signals directly from unlabeled data, thereby enabling scalable and cost-effective solutions for real-world agricultural systems \cite{Zhao2023CLA}.

Several SSL approaches have been specifically tailored for PDD. Fang et al. \cite{fang2021self} introduced the Cross-Iterative Kernel-Based Image Clustering System (CIKICS), which applied cross-iterative Kernel K-means to generate pseudo-labeled training sets. While this approach demonstrated promising results, its offline computations and slow processing speed constrained its scalability to large datasets. Similarly, Yang et al. \cite{Yang2020Self-supervised} proposed a feedback-driven mechanism for tomato disease detection, employing three localized leaf regions to optimize fine-grained categorization. Despite achieving high accuracy, the method was disease-specific, required balanced class distributions, and incurred significant computational overhead due to the reliance on multiple models.

Beyond domain-specific innovations, general SSL frameworks have significantly advanced representation learning in computer vision and, by extension, PDD. BYOL \cite{grill2020bootstrap} demonstrated the ability to learn without negative pairs by employing an online–target network strategy, where the target network was updated via momentum averaging. Although BYOL achieved state-of-the-art results on ImageNet and transfer tasks, its heavy reliance on data augmentation and lack of theoretical guarantees to prevent representational collapse posed challenges. Building on this, SimSiam \cite{chen2021exploring} simplified SSL further by removing the need for negative pairs, momentum encoders, and large batch sizes. Representation collapse was mitigated through a stop-gradient operation, allowing competitive ImageNet performance. However, its dependence on augmentations and the absence of strong theoretical foundations remain limiting factors.

The CIKICS framework was later extended with deep learning integration, combining Kernel K-means clustering with convolutional neural networks, ResNet50-based similarity metrics, and Siamese architectures \cite{fang2021self}. This enabled improved clustering of unlabeled plant disease images across both balanced and unbalanced datasets. Nevertheless, its iterative structure and high computational demand restricted its use in real-time agricultural monitoring. Parallel to this, Zhang et al. \cite{Zhang2021Detection} employed SimCLR for cassava leaf disease detection, showing strong classification performance but with a dependence on large batch sizes and augmentations that constrained generalizability. Bunyang et al. \cite{Bunyang2023Self-supervised} similarly leveraged SimCLR as a pre-trained backbone for PDD, though this required substantial computational and memory resources.

The adoption of Vision Transformers (ViTs) has further diversified SSL-based approaches in PDD. Zhang et al.~\cite{zhang2023information} introduced entropy masking for tea disease classification, but the approach risked discarding critical features during sampling. Yu et al.~\cite{yu2023mix} proposed a mixed-attention ViT for ultra-fine categorization of plant diseases, yet its high parameter count hindered efficient disease localization. Likewise, Meng et al.~\cite{Meng2023Known} exploited pre-trained ViTs to classify both known and unknown diseases but faced difficulties distinguishing visually similar classes, which amplified computational complexity and necessitated larger datasets.

More recent contributions include Monowar et al. \cite{monowar2022self}, who developed a self-supervised clustering method aimed at improving stability. Although effective, reliance on random sample selection and heavy augmentation reduced its generalizability. Fang et al. \cite{fang2021self} further refined the CIKICS framework by incorporating pre- and post-processing stages, enhancing clustering performance at the expense of higher computational costs and offline-only feasibility. Complementing these, Zhao et al. \cite{Zhao2023CLA} proposed Contrastive Learning for Leaf Disease Identification (CLA), which leveraged domain adaptation and pre-training to enhance generalization on limited and noisy labeled datasets. While CLA improved robustness, its accuracy was highly sensitive to data mix ratios, class distribution, and dataset size.

\subsection{Mamba Frameworks}

The limitations of CNNs and Transformers have driven the search for architectures balancing efficiency and representational power. CNNs excel at local feature extraction but suffer from limited receptive fields, while Transformers capture long-range dependencies at high computational cost. SSMs \cite{gu2021efficiently, gu2021combining} reframe sequence modeling through latent state transitions with linear-time complexity and strong long-range modeling capacity. Building on this, Mamba \cite{gu2023mamba} introduces input-dependent selective state updates and hardware-aware scanning, enabling dynamic adaptation to context. This design bridges efficiency and expressivity while avoiding the locality and gradient issues of earlier models.

The application of Mamba to vision tasks led to the development of the Vision Mamba \cite{zhu2024vision} backbone, which replaces self-attention with bidirectional state-space modules. Vision Mamba has been shown to deliver robust performance across classification, detection, and segmentation benchmarks. Building on this, PlainMamba \cite{yang2024plainmamba} introduced continuous two-dimensional scanning and direction-aware updates, simplifying the framework by avoiding hierarchical scaling. Other adaptations have focused on spectral and infrared imaging: HSIMamba \cite{yang2024hsimamba} and HSIDMamba \cite{liu2024hsidmamba} captured dependencies in hyperspectral data with efficiency and accuracy, while MiM-ISTD \cite{chen2024mim} nested Inner and Outer modules to enhance the detection of infrared small targets by jointly modeling global structures and fine-grained details.

Mamba has also shown strong adaptability in spatio-temporal and generative tasks. PoseMamba \cite{huang2025posemamba} integrated global and local bidirectional state modeling for improved monocular 3D pose estimation, while Motion Mamba \cite{zhang2024motion} embedded hierarchical temporal–spatial modules into diffusion models for efficient, realistic motion generation. Beyond this, MambaAD \cite{he2024mambaad} leveraged multi-scale locality-enhanced states with Hilbert-curve scanning for anomaly detection, and Samba \cite{bi2024samba} used severity-aware recurrent modules with EM-based recalibration to improve medical image grading under domain shifts. For low-level vision, VmambaIR \cite{shi2025vmambair} incorporated omni-selective scanning within a UNet backbone, boosting performance in deraining, deblurring, and super-resolution tasks.

In agriculture, Mamba-based approaches have demonstrated strong potential across diverse tasks. InsectMamba \cite{wang2025insectmamba} fused CNNs, self-attention, and SSMs via Mix-SSM blocks for robust insect pest recognition across datasets. GMamba \cite{zhang2024gmamba} leveraged Vision Mamba for grape leaf disease segmentation, improving boundary delineation in complex field settings. For disease detection, hybrid frameworks such as Mamba-YOLO-ML \cite{yuan2025mamba} integrated wavelet downsampling with Mamba modules into YOLO for mulberry leaf disease recognition, while YOLO-BSMamba \cite{liu2025yolo} embedded bidirectional state-space blocks into YOLOv8 to enhance tomato leaf disease detection under challenging conditions. VMamba-PlantDisease \cite{zhang2025vmamba} combined Vision Mamba with diffusion and transfer learning to address small-sample plant disease datasets. Extending beyond leaf-level analysis, UAV pipelines incorporating Mamba with attention and feature pyramid aggregation have been applied for pine wilt disease monitoring in forestry, highlighting Mamba’s scalability for large-scale environmental applications \cite{bai2025pine}. However, despite these advances, Mamba has not yet been applied to PDD within an SSL framework. This represents a clear research gap and directly motivates the development of our proposed ConMamba framework.

\subsection{Motivation}
The above-mentioned traditional convolutional networks and transformer-based frameworks either struggle to capture global and long-range dependencies or face significant computational complexity. Besides, these methods employ static loss functions that maximize similarity without aligning local and global features. Recently, Mamba frameworks have been widely applied to capture long-range dependencies in an image and have yielded promising results. However, traditional Mamba architectures are typically designed for supervised learning and depend on large, well-labeled datasets. This makes them less effective in environments where labeled data is scarce, such as in plant disease datasets. Without mechanisms for data imbalance problems, standard Mamba-based models might overfit the majority classes. This can lead to inadequate feature representations for underrepresented diseases, limiting the overall detection performance. Existing architectures may struggle to capture the invariant features necessary for robust PDD. Variations in lighting, angle, and plant condition require the model to learn features that are consistent across these changes, which is a gap that CL is well-suited to address.

Thus, to address these limitations, we propose the Contrastive Mamba (ConMamba) framework for PDD. It has the following advantages: it can harness large amounts of unlabeled data by learning invariant features from different augmentations of the same image. This is particularly useful in PDD, where obtaining labeled data is challenging. By training the model to distinguish between different augmented views of the same plant image (positive pairs) and different plant images (negative pairs), CL forces the model to focus on essential, disease-relevant features. CL can also help to balance the representation learning process by emphasizing the similarity between different views of the same disease condition, rather than relying solely on class labels that might be skewed. All these benefits will contribute to learning more discriminative and robust feature representations, which can improve PDD accuracy.

\section{Preliminaries} \label{sec:Preliminaries}

\subsection{State Space Model (SSM)}
SSMs mathematically describe continuous linear systems using structured linear differential equations. An SSM transforms an input sequence $x(t) \in \mathbb{R}^L$ into an output sequence $y(t) \in \mathbb{R}^L$ through an intermediate hidden state $h(t) \in \mathbb{R}^N$. This hidden state evolves according to the state transition matrix $A \in \mathbb{R}^{N \times N}$, influenced by input and output projection parameters $B \in \mathbb{R}^{N \times 1}$ and $C \in \mathbb{R}^{1 \times N}$:
\begin{equation}
    h'(t) = Ah(t) + Bx(t), \quad y(t) = Ch(t) + Dx(t).
\end{equation}
For discrete inputs, which are common in deep learning, discrete parameters $\bar{A}$ and $\bar{B}$ are defined using a time step $\Delta$:
\begin{equation}
    \bar{A} = \exp(\Delta A), \quad \bar{B} = (\Delta A)^{-1}(\exp(\Delta A)-I)\Delta B,
\end{equation}
resulting in a discrete model:
\begin{equation}
    h_t = \bar{A}h_{t-1} + \bar{B}x_t, \quad y_t = Ch_t.
\end{equation}
The S4 model \cite{gu2021efficiently} uses fixed parameters $A$, $B$, $C$, and $\Delta$ across all time steps, enhancing computational efficiency through global convolution but potentially limiting adaptability to specific inputs.

\subsection{Selective State Space Model}
The selective SSM \cite{gu2023mamba} introduces a time-variant mechanism by dynamically computing the parameters $B$, $C$, and $\Delta$ based on the input sequence via linear projection, while keeping $A$ fixed. 
This dynamic adaptation enhances sequence modeling but eliminates the feasibility of global convolution, requiring recurrent processing instead. 
Although effective for sequential data, this approach limits GPU parallelization, leading to slower performance. To mitigate this issue, the Mamba model integrates a hardware-optimized algorithm that enables efficient GPU utilization for recurrent computations. Designed specifically for one-dimensional sequential data, such as text and audio, Mamba provides a practical solution for tasks requiring adaptive sequence processing.

\subsection{Self-supervised Learning with Contrastive Learning}

CL is a prominent self-supervised representation learning paradigm, designed to differentiate between similar (positive) and dissimilar (negative) samples within a dataset, enabling the model to learn useful representations without requiring explicit labels \cite{hu2024comprehensive}. By strategically constructing pairs of data points, this technique brings similar instances closer while pushing dissimilar instances apart in the feature space, facilitating the capture of underlying semantic structures inherent to the data. This process relies heavily on the careful selection of positive and negative samples and the implementation of data augmentation strategies to generate informative contrasts \cite{chen2020simple}. Notably, CL frameworks such as SimCLR  \cite{chen2020simple}, MoCo \cite{he2020momentum}, and BYOL \cite{grill2020bootstrap} have demonstrated remarkable success across various domains, including computer vision, natural language processing, and speech recognition, achieving performance comparable to supervised learning approaches.

\section{Method}
\label{sec:method}
In this section, we introduce our proposed contrastive VM framework in details. First, we will give a detailed explanation of the advantages of VM for plant disease detection, and then introduce the contrastive VM step by step.

\begin{figure}[!t]
    \centering
    \subfigure[CNN]{
        \includegraphics[width=0.25\textwidth]{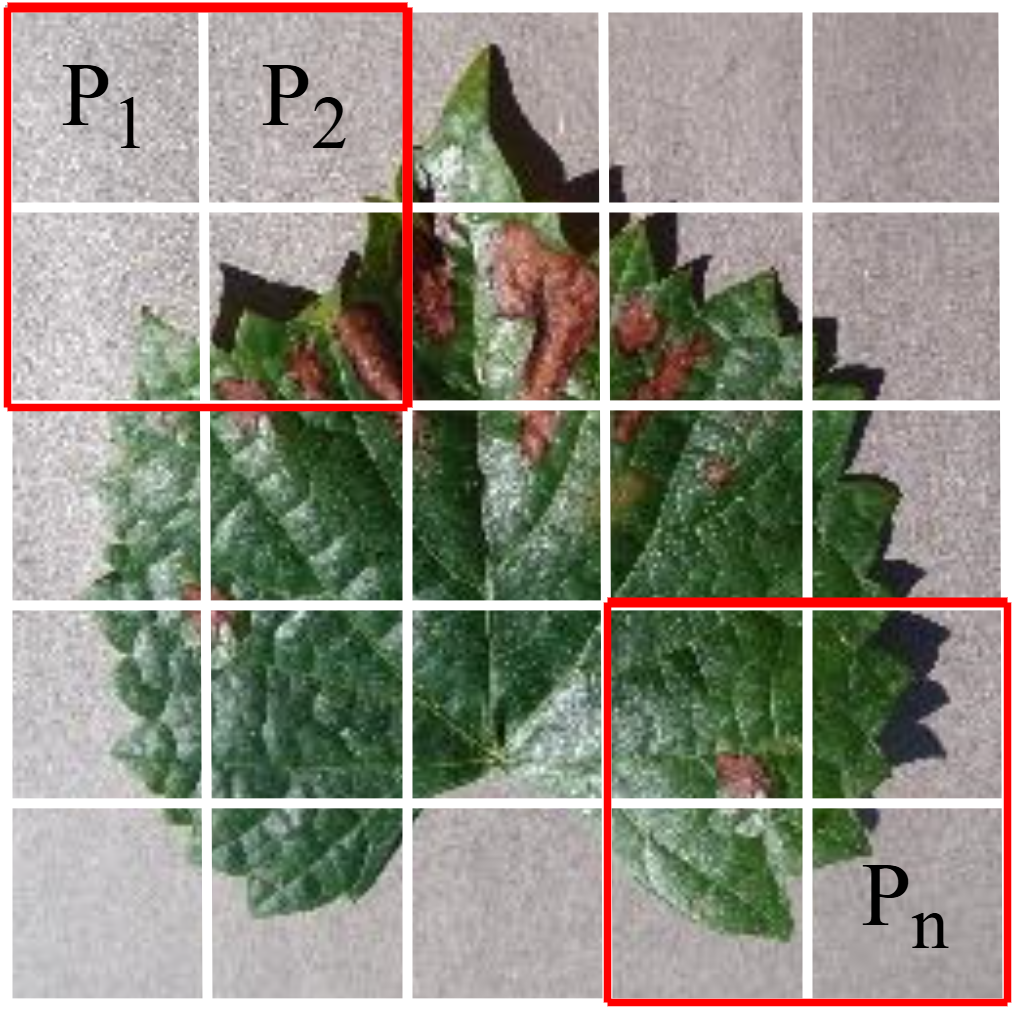}
    }
    \subfigure[ViT]{
        \includegraphics[width=0.25\textwidth]{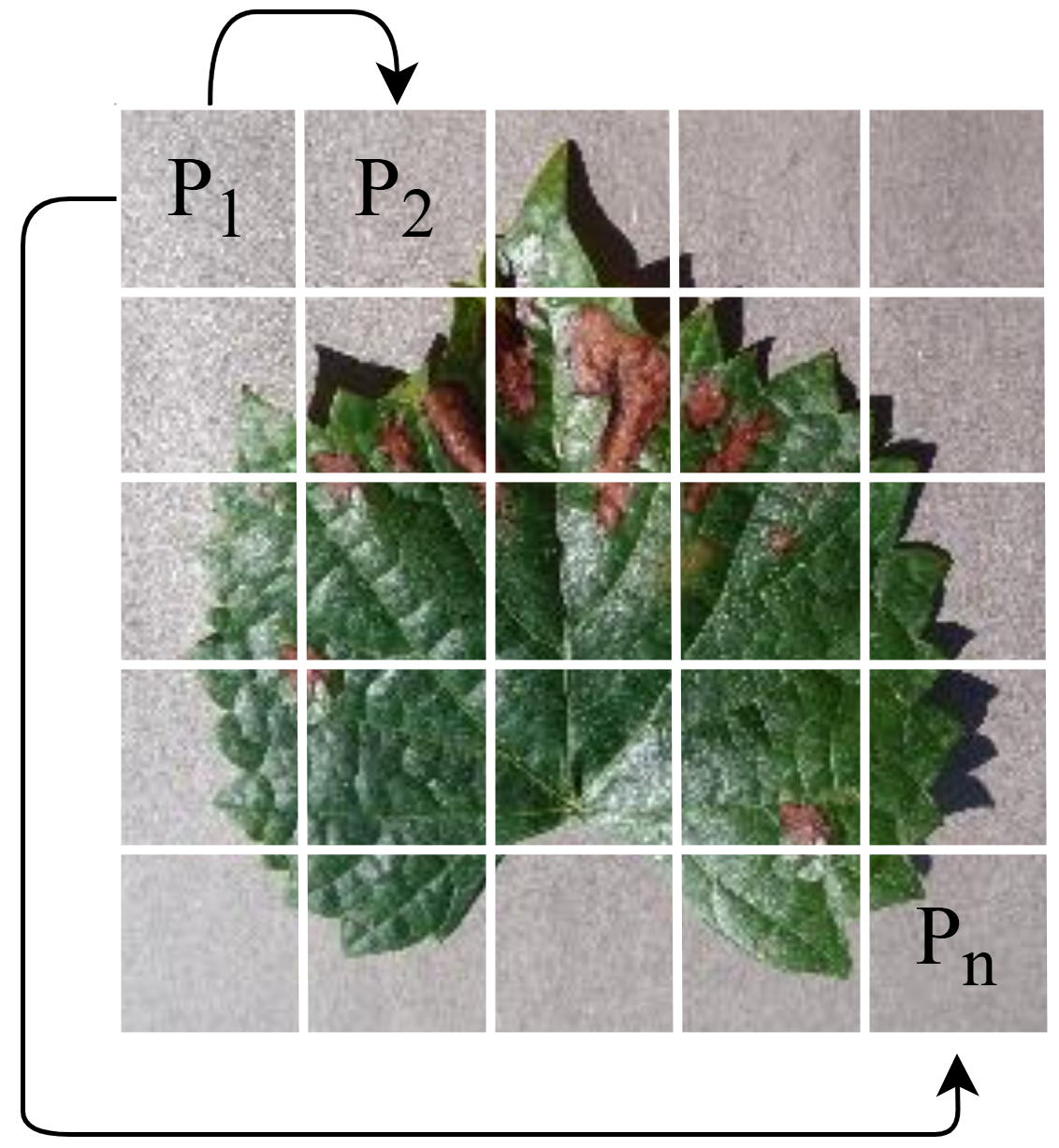}
    }
    \subfigure[Vision Mamba]{
        \includegraphics[width=0.25\textwidth]{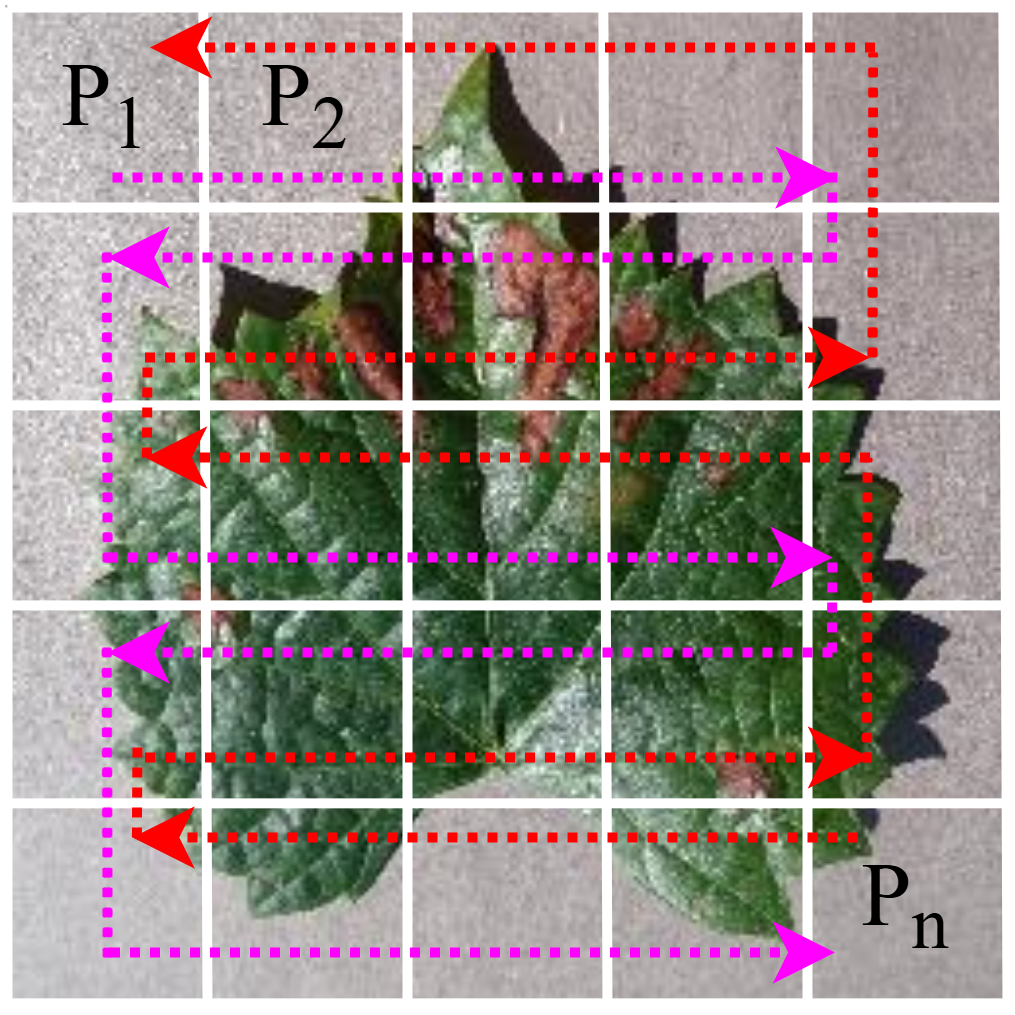}
    }
    \caption{Comparison of different architecture designs: (a) CNN, (b) ViT, and (c) Vision Mamba, emphasizing their capacities to capture short-range and long-range dependencies.}
    \label{fig:comparison architecture}
\end{figure}

\subsection{Vision Mamba: Rethinking Local-Global Context Modeling} \label{sec:vision mamba}
Current deep learning approaches, such as CNNs and Vision Transformers (ViTs), face challenges in capturing both local detailed information and global long-range dependencies effectiviely. CNNs utilize an inductive bias that emphasizes local connectivity, processing adjacent regions within an image using predefined kernel sizes, which is defined as:
\begin{equation}
    f(x)=\sigma(W*x+b),
\end{equation}
where $*$ denotes convolution, $W$ is the kernel, and $\sigma(\cdot)$ is a nonlinear activation. This operation effectively captures localized spatial fine detailed patterns, such as spatial relationships between nearby structures, as illustrated by regions P$_{1}$ and P$_{2}$ in Figure \ref{fig:comparison architecture} (a). However, CNNs struggle to integrate distant visual features due to their constrained receptive fields, limiting their ability to relate distant areas within an image, such as segment P$_{1}$ with P$_{n}$.
In contrast, ViTs eliminate these fixed inductive biases and treat all image patches equally. They capture global information through self-attention mechanisms and positional encoding to learn spatial relationships across an entire image, which is shown in Figure \ref{fig:comparison architecture} (b). The self-attention is defined as:
\begin{equation}
    \text{Attention}(Q,K,V)=\text{softmax}\left(\frac{QK^T}{\sqrt{d_k}}\right)V,
\end{equation}
where $Q,K,V$ represent query, key, and value embeddings respectively, and $d_k$ is the embedding dimension.
 This design enables ViTs to capture global dependencies effectively, making it possible to establish direct connections between widely separated image regions. However, due to this self-attention mechanism, ViTs require extensive data and computational resources to adequately learn local and global features.

To overcome these limitations, the VM introduces a bidirectional SSM~\cite{gu2023mamba} to integrate local and global contextual information efficiently. Unlike CNN and ViT, VM models visual sequences of image patches bidirectionally, combining localized inductive biases with global dependency modeling. Therefore, the vision mamba is able to learn local dependencies, allowing it to model relationships between adjacent segments (e.g., P$_{1}$ and P$_{2}$). At the same time, its efficient long-sequence modeling capabilities extend its receptive field across distant regions.

This balanced combination of local and long-range feature extraction makes VM particularly well-suited for PDD, where both local and global information are crucial for accurate disease detection. Figure \ref{fig:comparison architecture} (c) illustrates how VM's sequential processing allows it to capture short- and long-range dependencies more effectively than CNNs and ViTs alone.

\begin{figure}[!t]
    \centering
    \subfigure[Affected in a specific region]{%
        \includegraphics[width=0.25\textwidth]{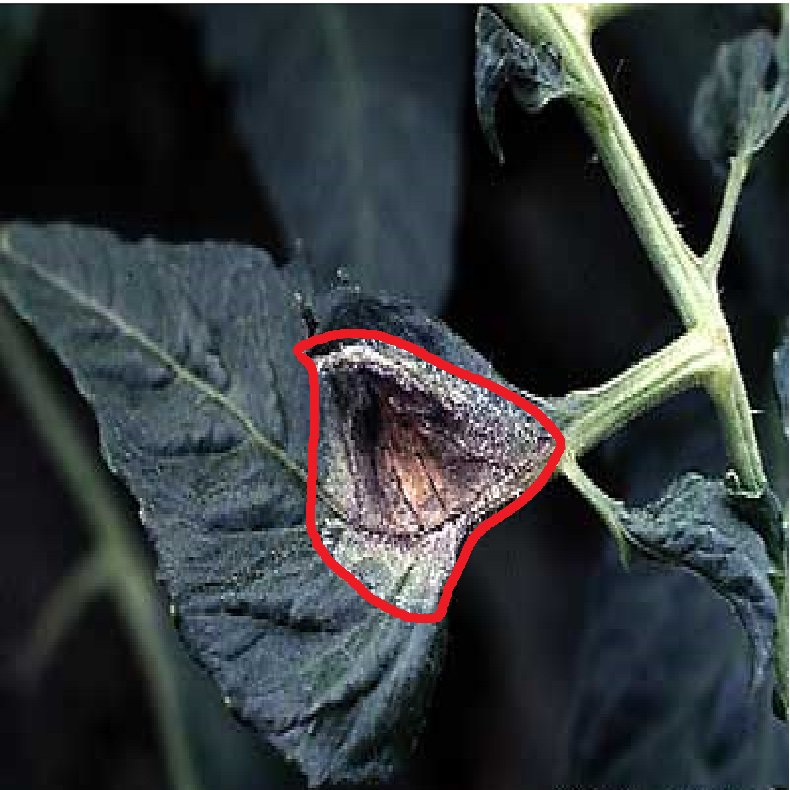}}
    \hspace{0.05\textwidth} 
    \subfigure[Affected across different areas]{%
        \includegraphics[width=0.25\textwidth]{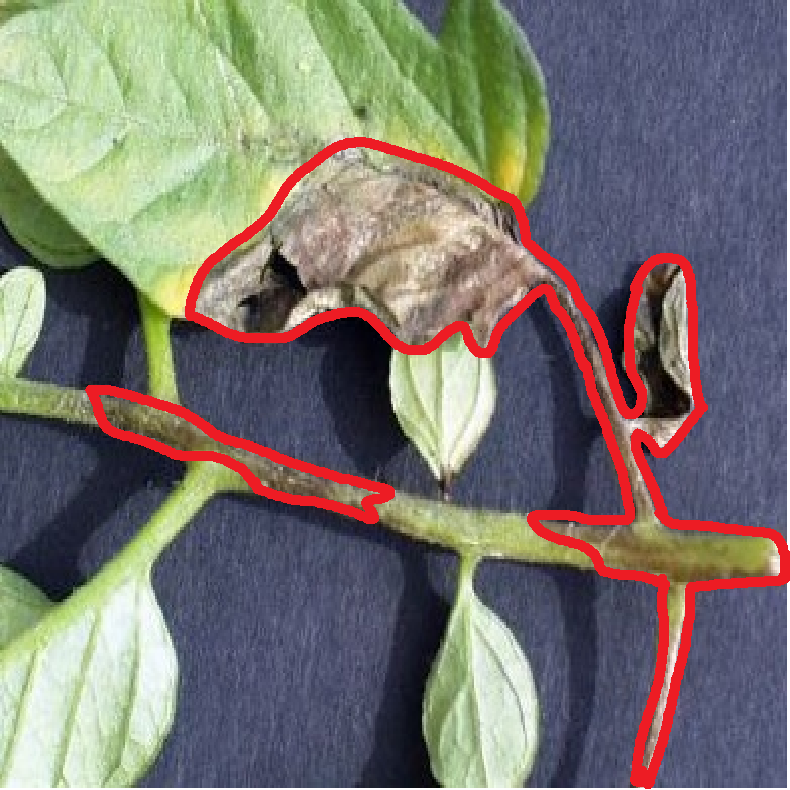}}
    \caption{Sample images of plant diseases: (a) affected in a specific region, and (b) affected across different areas, emphasizing the importance of local features and global, long-range dependencies for comprehensive plant disease detection.}
    \label{fig:sample local global}
\end{figure}

\subsection{Vision Mamba for Robust Plant Diseases} \label{sec:Vm for PDD}

Effective PDD demands the capability to identify subtle localized symptoms and recognize systemic disease patterns \cite{barbedo2016review} distributed across the entire plant. As shown in Figure \ref{fig:sample local global} (a), the plant leaf is affected in a specific region, where local features play a crucial role in accurate classification. In contrast, Figure \ref{fig:sample local global} (b) illustrates the effect of disease on various parts of the plant, highlighting the importance of global and long-range dependencies for identifying multiple affected regions. Thus, both local features and global, long-range dependencies are essential for comprehensive PDD.
Vision Mamba's bidirectional long-sequence modeling efficiently correlates distant regions within plant images. Suppose an input plant leaf image $X$ is segmented into patches $\{x_1,x_2, \cdots, x_M\}$. VM processes this patch sequence through its bidirectional SSM, represented as:\\
Forward Direction:
\begin{equation}
    h_t^{fwd}=\bar{A}h_{t-1}^{fwd}+\bar{B}x_t, ~~~~~~y_t^{fwd}=Ch_{t}^{fwd}
\end{equation}
Backward Direction:
\begin{equation}
    h_t^{bwd}=\bar{A}h_{t-1}^{bwd}+\bar{B}x_t, ~~~~~~y_t^{bwd}=Ch_{t}^{bwd}
\end{equation}
The final embedding of each patch $x_t$ integrates bidirectional features as:
\begin{equation}
   h_t=\text{Fuse}(y_t^{fwd},y_t^{bwd})
\end{equation}

This fusion of forward and backward contexts allows VM to robustly encode fine-grained localized features and complex global patterns, thereby significantly enhancing disease detection accuracy, even in scenarios with subtle or discattered symptoms.

\subsection{Contrastive Vision Mamba (ConMamba)}
The proposed Contrastive Vision Mamba (ConMamba) framework is specifically designed to effectively leverage self-supervised learning for robust plant disease detection. ConMamba introduces a dual-level contrastive learning approach that simultaneously optimizes embeddings for both local and global discriminability, uniquely integrated with a dynamic uncertainty-based weighting mechanism.

ComMamba consists of three key components:
\begin{itemize}
\item Contrastive Data Augmentation: We generate semantically consistent yet visually diverse augmented viewers from each original image, forming the positive and negative pairs necessary for contrastive learning.
\item Bidirectional Feature Representation with Vision Mamba: Leveraging Vision Mamba's novel bidirectional SSM, we transform augmented image views into highly discriminative embeddings. 
\item Dual-level Contrastive Optimization: we propose two complementary yet distinct contrastive losses, intra-class contrastive loss and inter-class contrastive loss, to optimize local robustness and enforce clear discriminative boundaries among embeddings from distinct disease classes.
\end{itemize}

\begin{figure}[t]
    \centering
    \includegraphics [width=\linewidth]{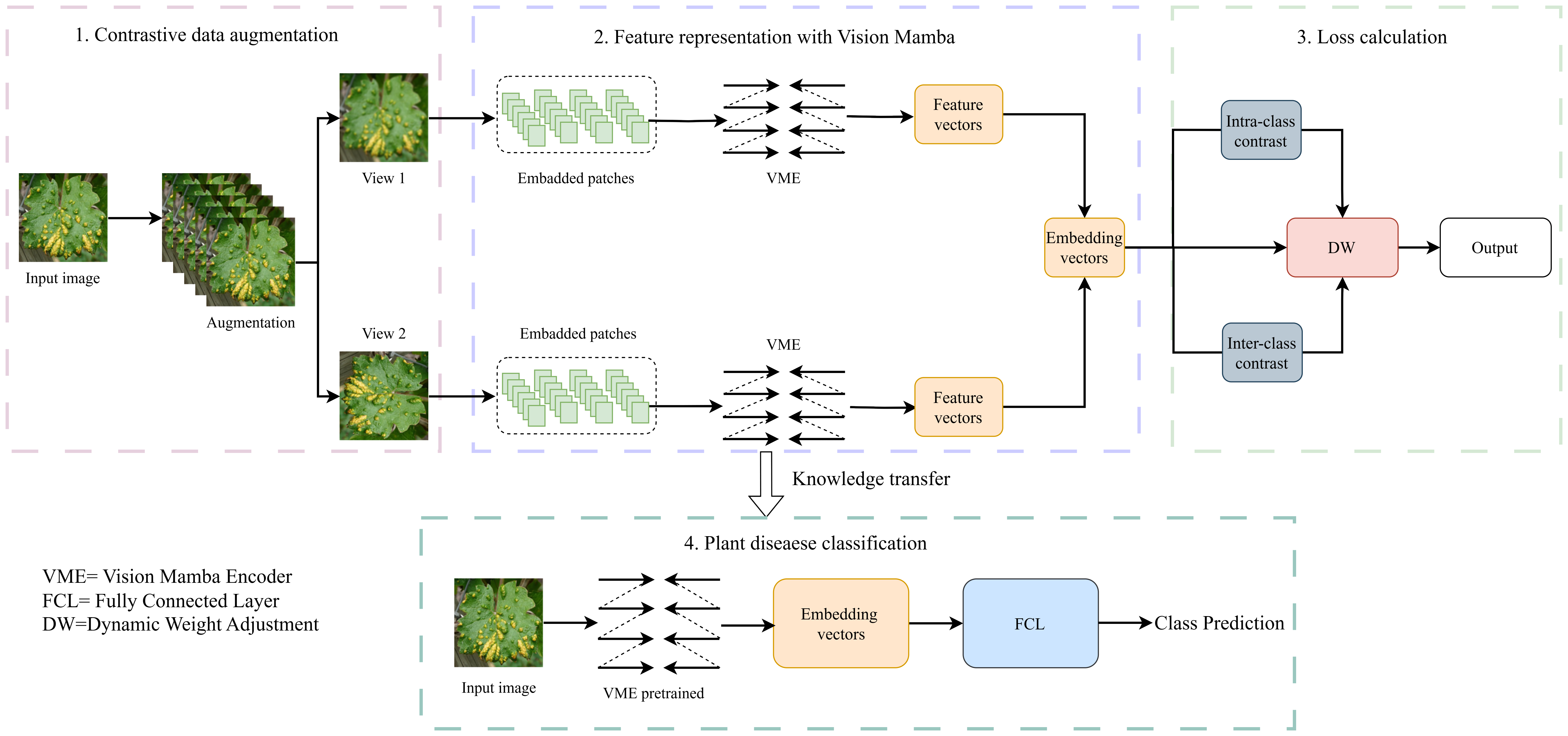}
    \caption{Schematic representation of the ConMamba framework. The framework begins with Stage 1 (Contrastive data augmentation): Data augmentation is applied to input images to generate two distinct augmented views for each input image. Stage 2 (Feature representation with Vision Mamba): Each augmented view undergoes patch embedding followed by the Vision Mamba Encoder (VME) to obtain meaningful bidirectional feature representations. Stage 3 (Loss calculation): A dual-level Contrastive loss with dynamic weight adjustment is employed to maximize local pairwise similarity (intra-class contrast) and global alignment (inter-class contrast). Finally, Stage 4 (Plant disease classification): The plant disease classification task adapts the learned representations for plant disease classification, utilizing the embedding vectors to produce class predictions.}
    \label{fig:01}
 \vspace{-8pt}
\end{figure}

The integration of these three components provides ConMamba with a robust framework to simultaneously capture fine-grained local variations and clear global discriminative structures, significantly enhancing representation quality for plant disease detection tasks. An overview of the proposed model is shown in Figure \ref{fig:01}.

\subsubsection{Contrastive Data Augmentation} \label{sec:Data Augmentation}
An important step in our Contrastive Mamba framework is the generation of meaningful contrastive samples by data augmentation. Specifically, each original plant disease image $X$ undergoes multiple augmentation operations $T(\cdot)$, resulting in different yet semantically consistent augmented views. Given an original image $X$, we generate augmented views $X_i=T_i(X)$, where each $T_i$ represents a random transformation, such as rotation, cropping, color jittering, or Gaussian noise addition. The augmented views can be defined as:
\begin{equation}
    \mathcal{X} = \{X_1, X_2, \cdots, X_n\}, ~~~~~~ X_i=T_i(X)
\end{equation}
These augmented views serve as positive pairs from the same original image for intra-contrastive learning, which will focus on local feature robustness. Simultaneously, augmented views from different original images form negative pairs, utilized inter-contrastive learning to ensure global feature discriminability. This formulation of augmented samples provided the foundational data required for subsequent loss calculation in Section 4.3.3 and facilitates robust, discriminative embedding learning within our Contrastive Vision Mamba framework.

\begin{figure}[!t]
    \centering
    \includegraphics[width=0.7\linewidth]{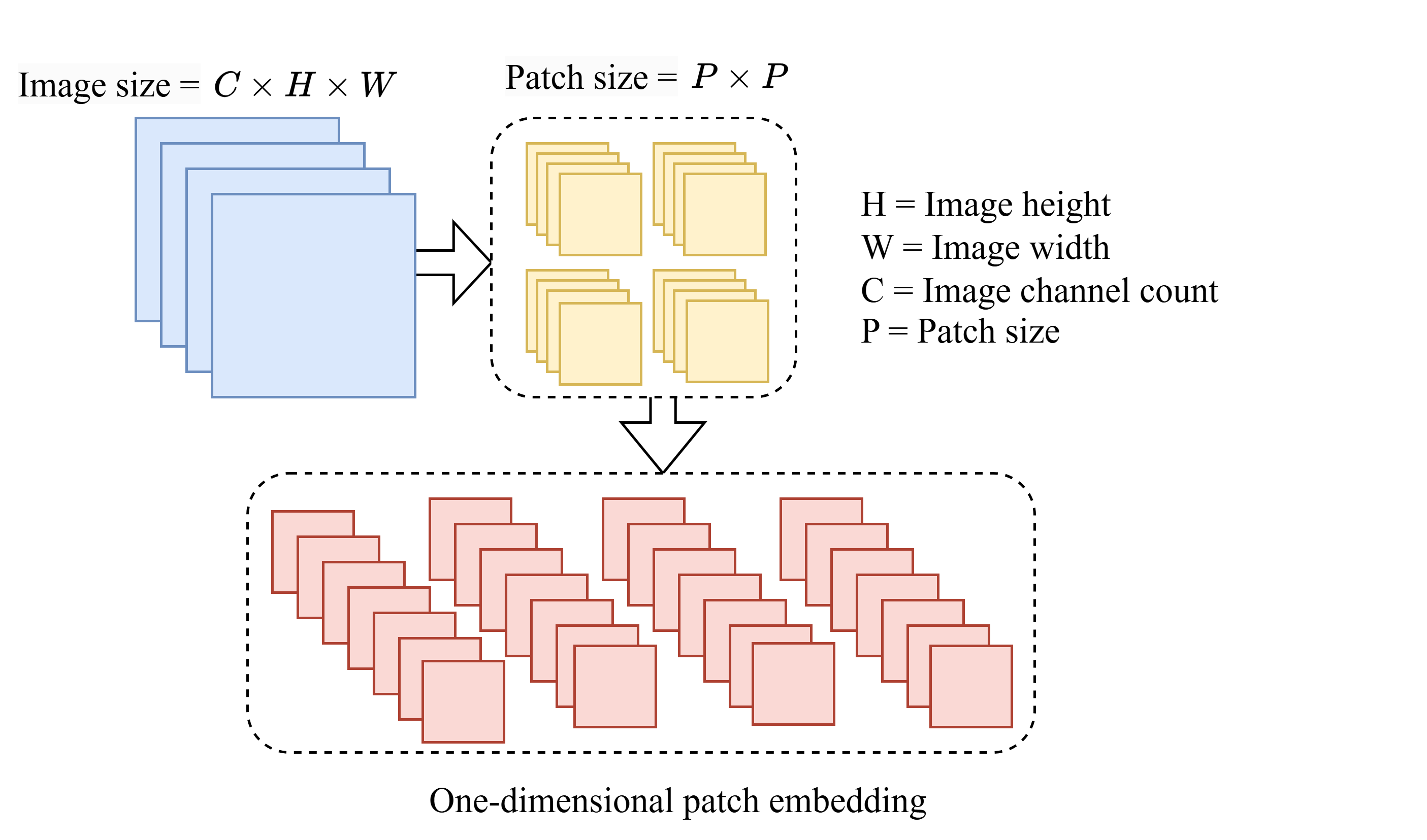}
    \caption{Illustration of the patch embedding process.}
    \label{fig:patch}
    \vspace{-8pt}
\end{figure}

\subsubsection{Feature Representation with Mamba}
After generating the augmented views as described in Section \ref{sec:Data Augmentation}, we transform each augmented image into discriminative embeddings using the Mamba encoder. Specifically, each augmented image $X_i\in \mathcal{X}_{aug}$ is first divided into a sequence of patches $\{x_i^1,x_i^2, \cdots, x_i^M\}$ as shown in Figure \ref{fig:patch}, where $M$ denotes the total number of patches. Each patch $x_i^j$ is embedded into a lower dimensional representation $h_i^j$ via the Mamba encoder:
\begin{equation}
    h_i^j = \text{MambaEncoder}(x_i^j)
\end{equation}
This step leverages the Mamba's encoder's capability to capture rich bidirectional contexts, enabling both local and global features to be robustly represented. These embedded representations $\{h_i^1, h_i^2, \cdots, h_i^M\}$ serves as inputs for the subsequent intra- and inter-contrastive loss computations in Section 4.3.3, providing structured, context-rich embeddings that significantly enhance the model's capability for accurate and robust plant disease detection.

\subsubsection{Dual-level Contrastive Learning and Dynamic Weight Adjustment}
\begin{figure}[!t]
    \centering
    \includegraphics[width=0.5\linewidth]{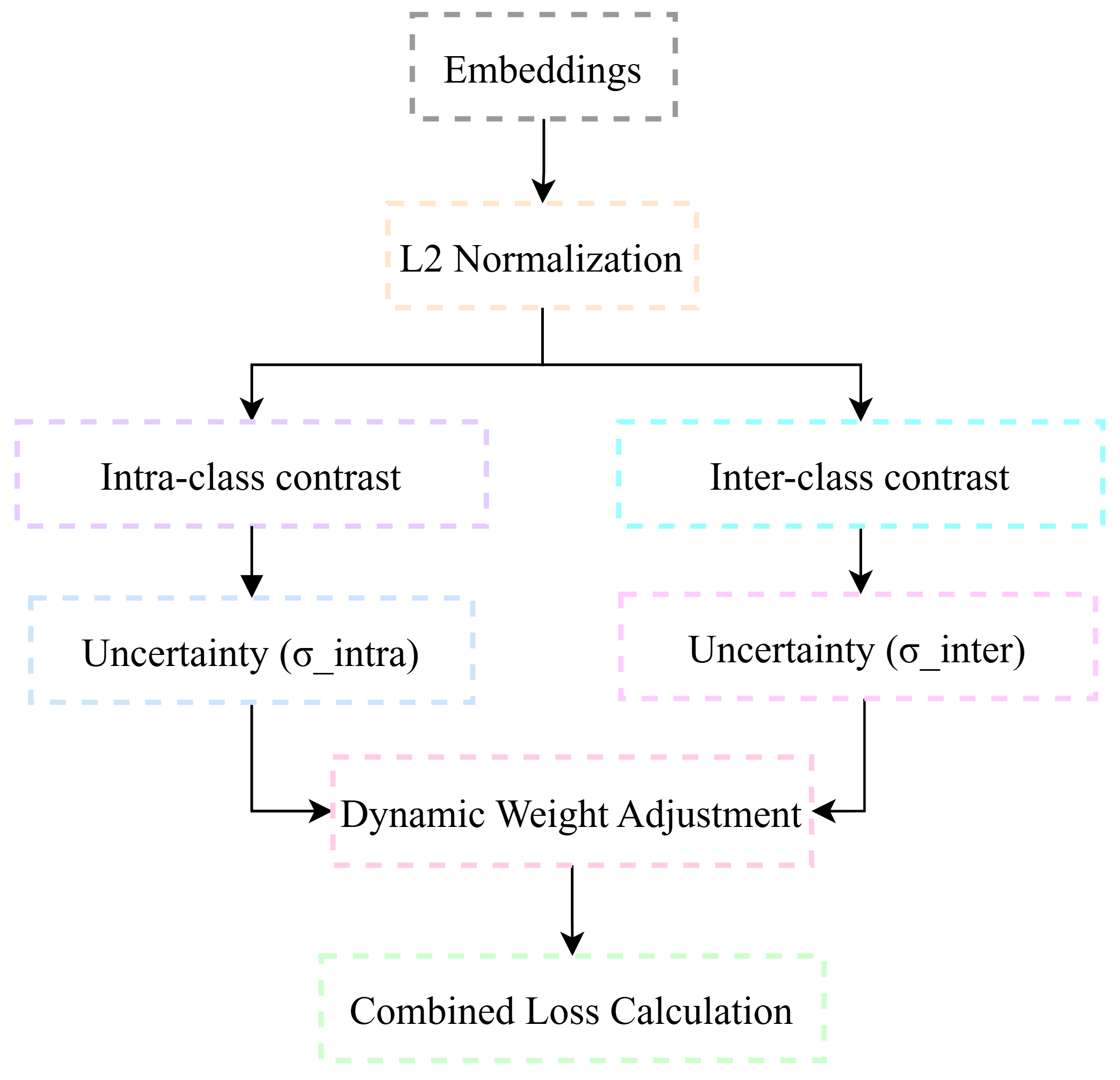}
    \caption{Dual-level Contrastive loss with dynamic weight adjustment.}
    \label{fig:04}
    \vspace{-10pt}
\end{figure}

In our Contrastive Mamba framework, the loss calculation is essential to effectively learning discriminative representations from plant disease images. To achieve a comprehensive embedding space, we propose a dual-level contrastive learning strategy comprising: Intra-Class Contrastive Learning and Inter-Class Contrastive Learning. The intra-class component enhances local feature robustness by maximizing the similarity between embeddings derived from different augmented viewers of the same image, while the inter-class component improves global discriminability by enforcing embeddings from different disease classes to be distinctly separated. Furthermore, we introduce a dynamic weight adjustment mechanism that adaptively combines these two losses to optimize overall performance. An overview of this dual-level contrastive loss with dynamic weight adjustment is shown in Figure \ref{fig:04}.

\begin{itemize} 
\item \textbf{Intra-Class Contrast:} Contrastive loss promotes local consistency by ensuring embeddings derived from different augmented views of the same image are closely aligned. It aims to maximize the mutual information (MI) between representations from positive pairs and minimize the MI between representations from negative pairs. Given two augmented views $T_i(X)$ and $T_j(X)$ of an original input $X$, their embeddings $h_i$ and $h_j$ form a positive pair. Embeddings from different images form negative pairs. Intra-class contrastive learning aims to maximize this mutual information between the augmented views of the same image, thus encouraging embeddings to ignore irrelevant variations(noise, transformations). This optimization corresponds to minimizing the following Normalized Temperature scaled Cross Entropy loss:
\begin{equation}
    \mathcal{L}_{intra} = -\frac{1}{N} \sum_{i=1}^N \log \frac{\exp(h_i \cdot h_j / \tau)}{\sum_{k \neq i} \exp(h_i \cdot h_k / \tau)},
\end{equation}
where $\tau$ represents the temperature parameter controlling the sharpness of the similarity distribution, and  $N$ is the number of positive pairs in the batch.

The Vision Mamba framework actually enhances intra-class contrastive learning. Because the vision mamba in this paper uses a Bidirectional State Space Model(SSM), which means it captures both forward and backward context within image patches. By encoding local context from both directions, Mamba gains a richer and more robust local understanding of each image patch, helping to maintain consistent representations across augmentations. As the intra-class contrastive loss pushes embeddings from augmentations close together, the bidirectional modeling reinforces this robustness because it provides a more stable local representation, less sensitive to irrelevant variations. Embeddings learned via Mamba become robust against subtle, local distortions, leading to improved local feature robustness, allowing the model to identify subtle disease symptoms more reliably in diseased plant images, despite slight variations in image appearance.

\begin{algorithm}[!t]
    \caption{Contrastive Vision Mamba Framework for Plant Disease Detection}
    \label{alg:contrastive_mamba}
    \KwIn{Dataset \(\mathcal{D} = \{(X_i,y_i)\}_{i=1}^{N}\), augmentation \(T(\cdot)\), Encoder \(\mathcal{F}_\theta\), epochs \(E\), batch size \(B\), margin \(m\), temperature \(\tau\), initial uncertainties \(\sigma_{intra}, \sigma_{inter}\), learning rate \(\eta\)}
    \KwOut{Trained encoder parameters \(\mathcal{F}_{\theta^*}\)}

    Initialize parameters \(\theta\) for encoder \(\mathcal{F}_\theta\)\;
    Initialize uncertainty parameters \(\sigma_{intra}, \sigma_{inter}\)\;

    \For{epoch \(=1,2,\dots,E\)}{
        \For{each batch \(\{(X_b,y_b)\}_{b=1}^{B}\)}{

            \tcp{Contrastive Data Augmentation}
            Generate two augmented views per image: \(X_b^1 = T_1(X_b), X_b^2 = T_2(X_b)\)\;

            \tcp{Feature Representation with Mamba}
            Compute embeddings \(h_b^1, h_b^2\) using Mamba encoder:
            \[
                h_b^1 = \mathcal{F}_\theta(X_b^1), \quad h_b^2 = \mathcal{F}_\theta(X_b^2)
            \]

            \tcp{Intra-Class Contrastive Loss (Local Robustness)}
            \[
                \mathcal{L}_{intra} = -\frac{1}{B}\sum_{b=1}^{B}\log\frac{\exp(h_b^1 \cdot h_b^2/\tau)}{\sum_{k\neq b}\exp(h_b^1 \cdot h_k^2/\tau)}
            \]

            \tcp{Inter-Class Contrastive Loss (Global Alignment)}
            Select positive embedding \(h_j^1\) from same class, negative embedding \(h_k^1\) from different class:
            \[
                \mathcal{L}_{inter} = \frac{1}{B}\sum_{b=1}^{B}\max\left(0,m - d(h_b^1,h_j^1) + d(h_b^1,h_k^1)\right)
            \]

            \tcp{Dynamic Weight Adjustment}
            \[
                \mathcal{L}_{total} = \frac{1}{2\sigma_{intra}^2}\mathcal{L}_{intra} + \frac{1}{2\sigma_{inter}^2}\mathcal{L}_{inter} + \log(\sigma_{intra}\sigma_{inter})
            \]

            \tcp{Parameter Update}
             $\theta \leftarrow \theta - \eta\nabla_\theta\mathcal{L}_{total}$\\
            $\sigma_{intra}\leftarrow\sigma_{intra}-\eta\nabla_{\sigma_{intra}}\mathcal{L}_{total},\quad\sigma_{inter}\leftarrow\sigma_{inter}-\eta\nabla_{\sigma_{inter}}\mathcal{L}_{total}$
                
        }
    }
    \Return Train encoder \(\mathcal{F}_{\theta^*}\)\;
\end{algorithm}

\item \textbf{Inter-Class Contrast:}
To ensure global discriminative capability, the inter-class contrastive loss is designed to minimize mutual information between embeddings from different classes, increasing the separability across distinct disease categories. Specifically, this loss enforces a clear embedding space boundary between different disease classes. Embeddings from images of the same class are encouraged to cluster tightly together, while embeddings from distinct classes are pushed apart by at least a margin $m$. To achieve this, a margin-based contrastive formulation is defined as:
\vspace{-6pt}

\begin{equation}
\mathcal{L}_{inter}= \frac{1}{N} \sum_{i=1}^N \max \left(0, m - d(h_i, h_j) + d(h_i, h_k) \right),
\end{equation}
where $d(h_i, h_j)$ measures similarity between embeddings of the same class, $d(h_i, h_k)$ measures similarity between embeddings of different classes, and $m$ is a margin ensuring negative pairs remain sufficiently distant.
This loss enhances class-level structure, improving the separability of learned features in the embedding space.

As mentioned earlier, we use the mamba framework that leverages a selective bidirectional context via its state-space model, modeling global interactions between patches across the entire image. This long-range dependency modeling means Mamba encodes more global meaningful features, connecting subtle disease indicators across distant image regions. Consequently, Mamba enhances the effectiveness of inter-class contrastive learning because it generates consistent global embeddings, making it easier to separate embeddings of different classes clearly. Global feature separability significantly enhances disease classification performance reducing confusion between classes that may have similar local symptoms but distinct global distributions of patterns. Therefore, these settings leads to embeddings with improved class discriminability, enhancing the accuracy and generalization of the model across varied disease types.

\item \textbf{Dynamic Weight Adjustment:}
To optimally combine the intra-class contrastive and inter-class contrastive losses, we introduce a dynamic weight adjustment mechanism based on uncertainty modeling. This approach addresses the challenge of effectively balancing the contributions of these two distinct loss functions during the training process. Specifically, we assume that each loss function follows a Gaussian likelihood with uncertainty parameters $\sigma_{intra}$ and $\sigma_{inter}$ representing the inherent noise or uncertainty associated with the intra- and inter- class contrastive tasks, respectively.

The likelihood functions can be defined as Gaussian distributions:
{\small
\begin{equation}
p(\mathcal{L}_{\text{intra}}|\sigma_{\text{intra}}) \propto \frac{1}{\sigma_{\text{intra}}}\exp\!\left(-\frac{\mathcal{L}_{\text{intra}}}{2\sigma_{\text{intra}}^{2}}\right),
\quad
p(\mathcal{L}_{\text{inter}}|\sigma_{\text{inter}}) \propto \frac{1}{\sigma_{\text{inter}}}\exp\!\left(-\frac{\mathcal{L}_{\text{inter}}}{2\sigma_{\text{inter}}^{2}}\right).
\end{equation}
}

From a Bayesian perspective, maximizing these likelihoods with respect to the uncertainty parameters is equivalent to minimizing the negative log-likelihood(NLL). The NLL for the combined loss can be defined as: 
\begin{equation}
-\text{log}p(\mathcal{L}_{intra},\mathcal{L}_{inter}|\sigma_{intra},\sigma_{inter)})=-\text{log}[p(\mathcal{L}_{intra}|\sigma_{intra})p(\mathcal{L}_{inter}|\sigma_{inter})]
\end{equation}

By expanding and simplifying this expression, our dynamic weight adjustment loss is given by
\begin{equation}
    \begin{aligned}
    &\mathcal{L}_{total} = \frac{1}{2 \sigma_{intra}^2} \mathcal{L}_{intra} 
    + \frac{1}{2 \sigma_{inter}^2} \mathcal{L}_{inter}
    + \log(\sigma_{intra} \cdot \sigma_{inter}),\\
    &\mathcal{L}_{final} = \frac{1}{N} \sum_{i=1}^N \mathcal{L}_{total}^{(i)},
\end{aligned}
\end{equation}
where the terms $\frac{1}{2 \sigma_{intra}^2}$ and $\frac{1}{2 \sigma_{inter}^2}$ act as adaptive weight factors, dynamically balancing intra-class contrastive and inter-class contrastive losses based on the learned uncertainty.

This ensures that the model places higher weight on the loss with lower uncertainty (higher confidence) and lower weight on the more uncertain loss. The additional term $\text{log}(\sigma_{intra},\sigma_{inter})$ acts as a regularizer, penalizing the uncertainty terms from becoming excessively large. Without this logarithmic regularization term, the uncertainty parameters could increase indefinitely, minimizing the contribution of both lossses and leading to ineffective learning. By introducing this $\text{log}(\cdot)$ term, we encourage the model to find a meaningful trade-off between minimizing individual losses and maintaining manageable uncertainty levels. Based on this analysis, we know that this dynamic uncertainty-based weighting allows the network to adaptively and continuously prioritize either local alignment(intra-class contrastive) or global alignment (inter-class contrastive) based on which aspect requires more attention during training. Therefore, the learned embeddings are highly discriminative, robust, and well-suited for downstream tasks such as PDD classification.

\end{itemize}

\subsection{Plant Disease Classification}
The Plant disease classification task, illustrated in Figure \ref{fig:01} (Step 4), processes input images using the pre-trained VME. Trained during the pretext task, the VME extracts meaningful and enriched features, encapsulating them in an embedding vector. This vector serves as a compact yet comprehensive representation of the images, preserving both local and global relationships essential for classification.
The embedding vector is then fed into a Fully Connected Layer (FCL), which maps the extracted features to their corresponding output classes. Finally, the FCL produces the predicted class labels, completing the downstream task.

The proposed Contrastive Vision Mamba (ConMamba) for plant disease detection is summarized in Algorithm 1.

\section{Experiments}\label{sec:experiment}
In this section, we carry out experiments on three publicly available datasets: PlantVillage~\cite{al2024plant}, PlantDoc~\cite{Singh2020PlantDoc}, and Citrus~\cite{rauf2019citrus}, for the plant disease classification task, which serves both to demonstrate the efficacy of the proposed ConMamba model and validate the claims mentioned in the previous sections. Numerical and visual results are carried out successively to evaluate the performance of the proposed model and the quality of the learned features, with comparisons to state-of-the-art methods across various evaluation metrics. Ablation studies were also conducted to evaluate the contribution of encoders and each loss component in enhancing the discrimination between classes across the datasets.

\subsection{Experimental Setup}

We trained and evaluated ConMamba using a distributed setup across 10 NVIDIA GeForce RTX 2080 Ti GPUs (each with 11\,GB VRAM), leveraging PyTorch’s Distributed Data Parallel (DDP) with the NCCL backend for efficient multi-GPU synchronization. The experiments were executed under CUDA 11.8 with driver version 555.42.06 on Ubuntu 20.04. The model was implemented in Python 3.10.13 using PyTorch 2.1.1, torchvision 0.16.1, and torchaudio 2.1.1. We employed the Mamba selective state-space model package 2.2.3 to enable bidirectional state-space modeling in the Vision Mamba encoder.

\begin{table}[!t]
\centering
\scriptsize
\caption{Performance of ConMamba on the three datasets. Evaluation metrics include Accuracy (\%), Precision (\%), Recall (\%), and F1-score (\%).}
\vspace{-3pt}
\label{tab:combined}
\resizebox{0.6\linewidth}{!}{
\begin{tblr}{
  cells = {c},
  hline{1-2,5} = {-}{},
}
\textbf{Dataset} & \textbf{Accuracy} & \textbf{Precision} & \textbf{Recall} & \textbf{F1-score} \\
PlantVillage \cite{al2024plant}     & 98.62             & 96.74           & 97.59              & 97.38             \\
PlantDoc \cite{Singh2020PlantDoc}   & 94.29             & 93.88           & 93.87              & 93.97             \\
Citrus \cite{rauf2019citrus}         & 91.38             & 91.74           & 90.51              & 91.66             
\end{tblr}}
\vspace{-3pt}
\end{table}

\subsection{Datasets}

Three datasets, namely PlantVillage, PlantDoc, and Citrus, are used in the experiments, and each of them offers unique features and challenges that provide a proper foundation for assessing the model's performance across a variety of plant species, diseases, and environmental conditions. A detailed overview of each dataset is presented in the following subsections.

\noindent \textbf{PlantVillage.}
The PlantVillage \cite{al2024plant} is one of the most extensively utilized and publicly accessible resources for PDD research. Developed by Penn State University, it consists of approximately 54,306 high-quality images spanning 14 plant species.
It is categorized into 38 classes, including 26 disease classes and 12 healthy classes. The disease categories include 4 bacterial diseases, 17 fungal diseases, 2 viral diseases, 1 mite-induced disease, and 2 mould-related diseases, offering a rich resource for understanding diverse plant health challenges.

\noindent \textbf{PlantDoc.}
The PlantDoc~\cite{Singh2020PlantDoc} dataset consists of 2,598 RGB images depicting leaf diseases across 13 plant species. These images are categorized into 28 distinct disease types and are taken directly from natural environments, incorporating real-world conditions such as varying lighting, complex backgrounds, and diverse leaf orientations. This natural diversity enhances the dataset's complexity, enabling models to learn robust features for real-world disease detection.

\noindent \textbf{Citrus.}
The Citrus~\cite{rauf2019citrus} dataset was developed to support research in PDD and classification. It comprises 759 high-resolution images of healthy and diseased citrus fruits and leaves, specifically targeting five major diseases: Black Spot, Canker, Scab, Greening, and Melanose. 
For citrus fruits, the dataset includes 150 images categorized into Black Spot, Canker, Greening, Scab, and Healthy. Citrus leaves contain 609 images, classified into Black Spot, Canker, Greening, Melanose, and Healthy. All images were collected from  Sargodha, Pakistan, a tropical agricultural region. Our research specifically focuses on the citrus leaf images from this dataset.

\subsection{Evaluation Metrics}
Four metrics, accuracy \cite{monowar2022self}, precision \cite{wang2024classification}, recall \cite{cao2025small}, and F1-score \cite{al2023deep}, are used to evaluate the performance of the proposed method and that of the benchmarks. These metrics provide a comprehensive evaluation of the model’s predictive capabilities, highlighting different aspects of its performance, such as overall correctness, the ability to avoid false positives, and the ability to capture true positives.

\subsection{Results and Analysis}

\subsubsection{Quantitative Performance}
The overall performance of the proposed ConMamba framework across the three datasets is summarized in Table~\ref{tab:combined}. ConMamba achieves its highest performance on the PlantVillage dataset, followed by PlantDoc and Citrus. This performance trend reflects the varying levels of dataset complexity.
The PlantVillage dataset is relatively simple, as it was collected under controlled conditions, featuring a plain background and a single, centrally positioned leaf in each image. In contrast, the PlantDoc dataset presents greater challenges, as it was captured in natural environments, where images may contain multiple leaves with varying orientations and off-centre placements. 
The Citrus dataset is the most challenging, as it includes diseased leaves, with some diseases exhibiting similar visual symptoms, making classification more difficult. Despite these challenges, the proposed method consistently achieves over 90\%  accuracy across all evaluation metrics, demonstrating its robustness and strong potential for accurate PDD.

\begin{table}[!t]
\scriptsize
\centering
\caption{Performance comparison of ConMamba with existing baseline methods on the test datasets. Accuracy, Precision, Recall, and F1-score are used as the evaluation metrics, and the top-performing result is highlighted in bold.}
\label{tab:comparison}
\begin{tblr}{
  row{odd} = {c},
  row{4} = {c},
  row{6} = {c},
  row{8} = {c},
  row{10} = {c},
  row{12} = {c},
  row{14} = {c},
  row{16} = {c},
  row{18} = {c},
  row{20} = {c},
  cell{2}{1} = {r=5}{},
  cell{2}{2} = {c},
  cell{2}{3} = {c},
  cell{2}{4} = {c},
  cell{2}{5} = {c},
  cell{2}{6} = {c},
  cell{7}{1} = {r=5}{},
  cell{12}{1} = {r=5}{},
  cell{17}{1} = {r=2}{},
  cell{19}{1} = {r=2}{},
  hline{1-2} = {-}{},
  hline{7,12,17,19,21} = {1-6}{},
}
\textbf{Dataset} & \textbf{Method}   & \textbf{Accuracy} & \textbf{Precision} & \textbf{Recall} & \textbf{F1-score} \\
PlantVillage \cite{al2024plant}     & Clustering \cite{monowar2022self}        & 88.90              & -                  & -               & -                 \\
                 & CIKICS \cite{fang2021self}            & 89.10             & 89.50               & 83.50            & 79.70              \\
                 & CLA \cite{Zhao2023CLA}              & 90.52             & 90.64              & 90.40            & 90.58             \\
                 & CAE \& CNN \cite{bedi2021plant}         & 98.38             & 96.21              & 97.10           & 96.65             \\
                 & \textbf{ConMamba} & \textbf{98.62}    & \textbf{96.74}     & \textbf{97.59}  & \textbf{97.38}    \\
PlantDoc \cite{Singh2020PlantDoc}         & InceptionResNetV2 \cite{Singh2020PlantDoc} & 70.53             & -                  & -               & 70.00             \\
                 & NASNetMobile \cite{enkvetchakul2022stacking}     & 76.48             & -                  & -               & -                 \\
                 & MobileNetV3 \cite{nguyen2023effective}       & 83.00             & 82.28              & 82.52           & 82.77             \\
                 & CNN \cite{chung2024addressing}              & 87.42             & 86.73              & 86.27           & 86.48             \\
                 & \textbf{ConMamba} & \textbf{94.29}    & \textbf{93.88}     & \textbf{93.87}  & \textbf{93.97}    \\
Citrus \cite{rauf2019citrus}            & CIKICS \cite{fang2021self}           & 64.10             & 52.10               & 42.20            & 46.60              \\
                 & SimSiam \cite{chen2021exploring}          & 87.50             & -                  & -               & -                 \\
                 & BYOL \cite{grill2020bootstrap}             & 88.90             & 90.52              & 87.93           & 87.76             \\
                 & Clustering \cite{monowar2022self}       & 89.30             & 88.69              & 88.85           & 88.77             \\
                 & \textbf{ConMamba} & \textbf{91.38}    & \textbf{91.74}     & \textbf{90.51}  & \textbf{91.66}    \\
Tea Sickness \cite{shashwatwork_tea_leafs}              & IEM-ViT \cite{zhang2023information}           & 93.78             & 93.67              & 93.80           & 93.64             \\
                 & \textbf{ConMamba} & \textbf{95.27}    & \textbf{94.89}     & \textbf{94.75}  & \textbf{94.82}    \\
Apple Foliar \cite{thapa2020plant}             & Mix-ViT \cite{yu2023mix}           & 97.36             & 96.91              & 97.14           & 97.10             \\
                 & \textbf{ConMamba} & \textbf{98.05}    & \textbf{97.55}     & \textbf{97.62}  & \textbf{97.57}    
\end{tblr}
\end{table}

To validate the effectiveness of the proposed ConMamba framework, we conducted comprehensive comparisons with state-of-the-art (SOTA) self-supervised methods, including CAE \& CNN ~\cite{bedi2021plant}, CLA ~\cite{Zhao2023CLA}, CNN ~\cite{chung2024addressing}, MobileNetV3 \cite{nguyen2023effective}, Clustering ~\cite{monowar2022self}, and BYOL ~\cite{grill2020bootstrap}. These comparisons highlight the robustness and generalizability of our method across three datasets, ranging from controlled to real-world agricultural environments. Table \ref{tab:comparison} reports a comparative evaluation of the proposed ConMamba framework against baseline methods across three test datasets.
On PlantVillage, a curated dataset with clean backgrounds and minimal interclass noise, ConMamba achieves an accuracy of 98.62\%. This marginally surpasses CAE \& CNN \cite{bedi2021plant} (98.38\%) and outperforms self-supervised methods such as CLA \cite{Zhao2023CLA} (90.52\%) and CIKICS \cite{fang2021self} (89.10\%). Additionally, ConMamba yields the highest precision (96.74\%), recall (97.59\%), and F1-score (97.38\%). These reflect strong intraclass consistency and generalizability even in less complex visual domains.

\begin{figure}[!t]
    \centering
    \begin{tabular}{ccc:cc:cc} 

        \rotatebox{90}{\scriptsize \text{~~~~Original}} &  
        \includegraphics[width=0.14\textwidth]{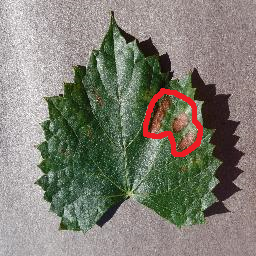} &
        \includegraphics[width=0.14\textwidth]{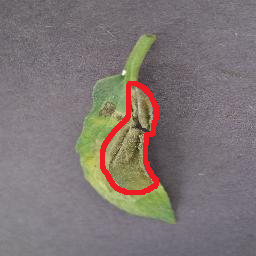} & 
        \includegraphics[width=0.14\textwidth]{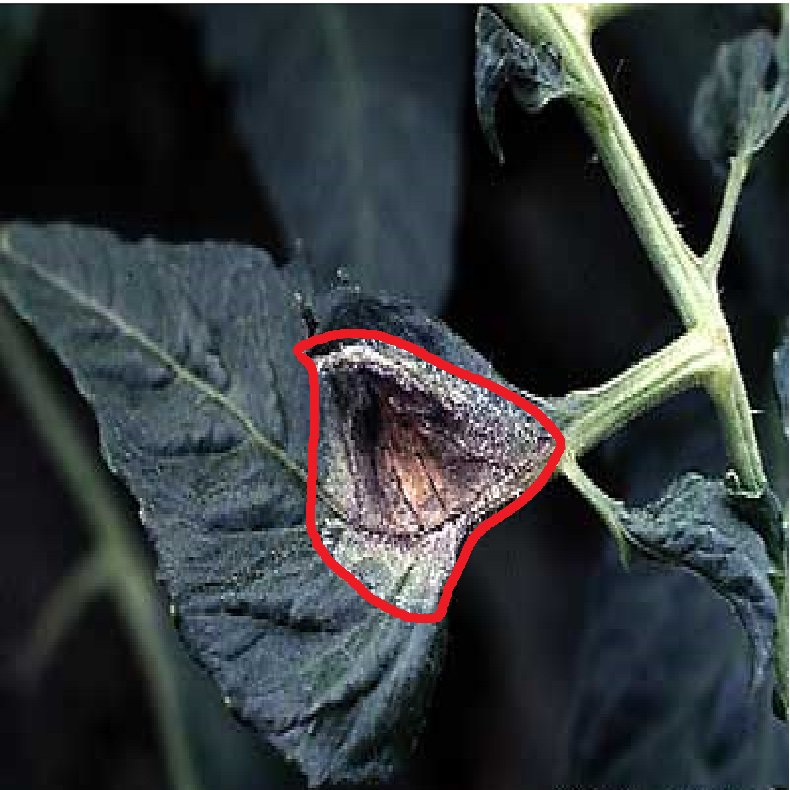} &
        \includegraphics[width=0.14\textwidth]{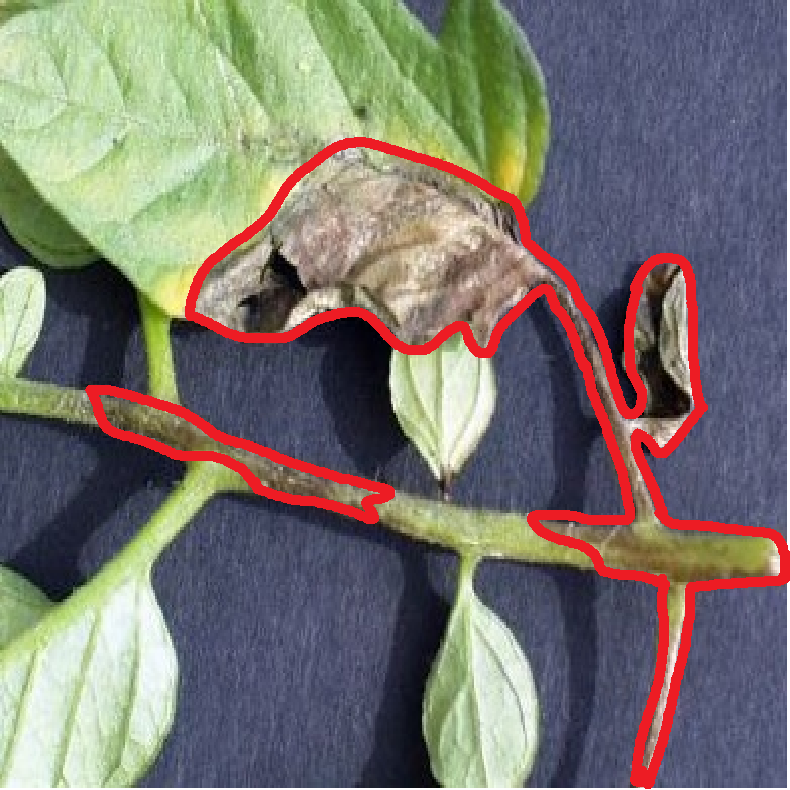} & 
        \includegraphics[width=0.14\textwidth]{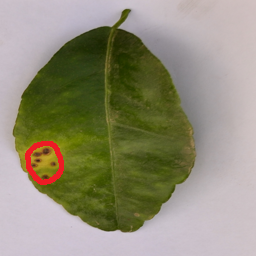} &
        \includegraphics[width=0.14\textwidth]{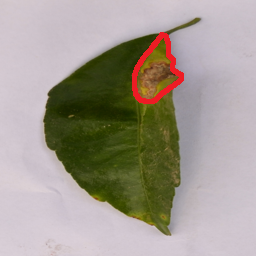} \\ 

        \rotatebox{90}{\scriptsize \text{~~~Second-best}} &  
        \includegraphics[width=0.14\textwidth]{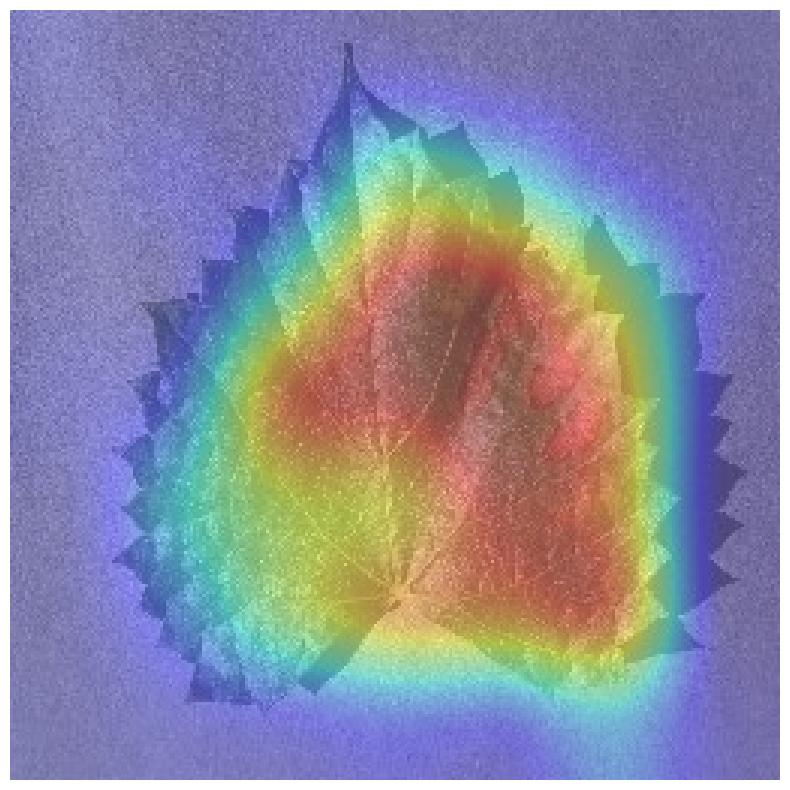} &
        \includegraphics[width=0.14\textwidth]{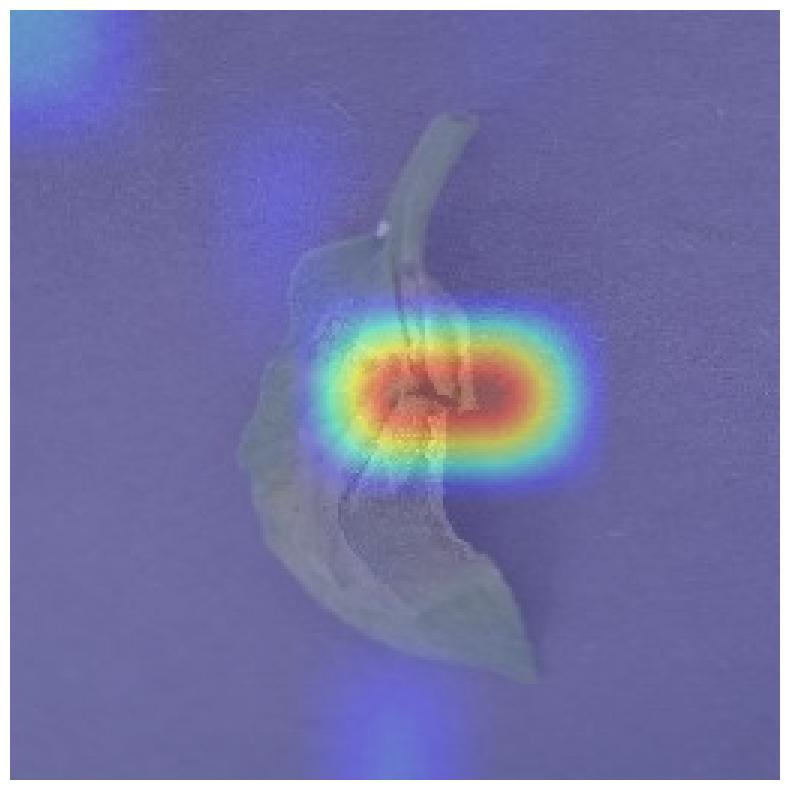} & 
        \includegraphics[width=0.14\textwidth]{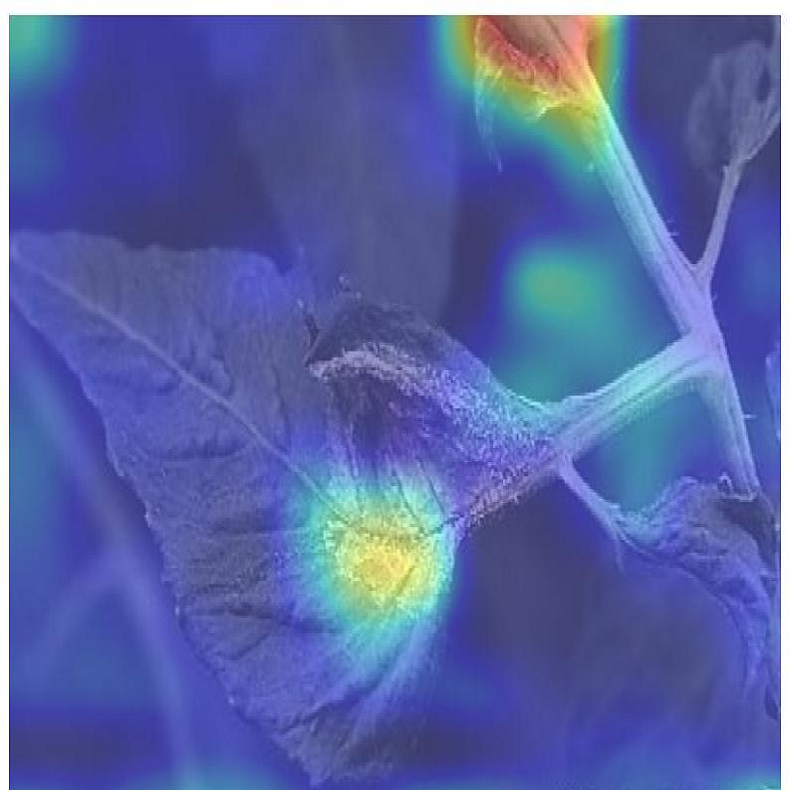} &
        \includegraphics[width=0.14\textwidth]{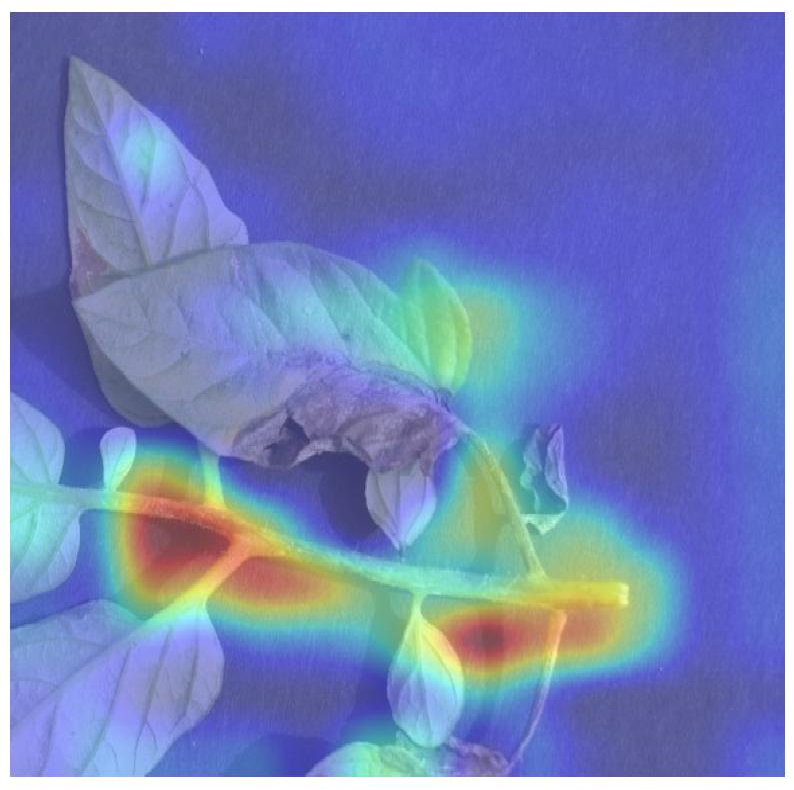} & 
        \includegraphics[width=0.14\textwidth]{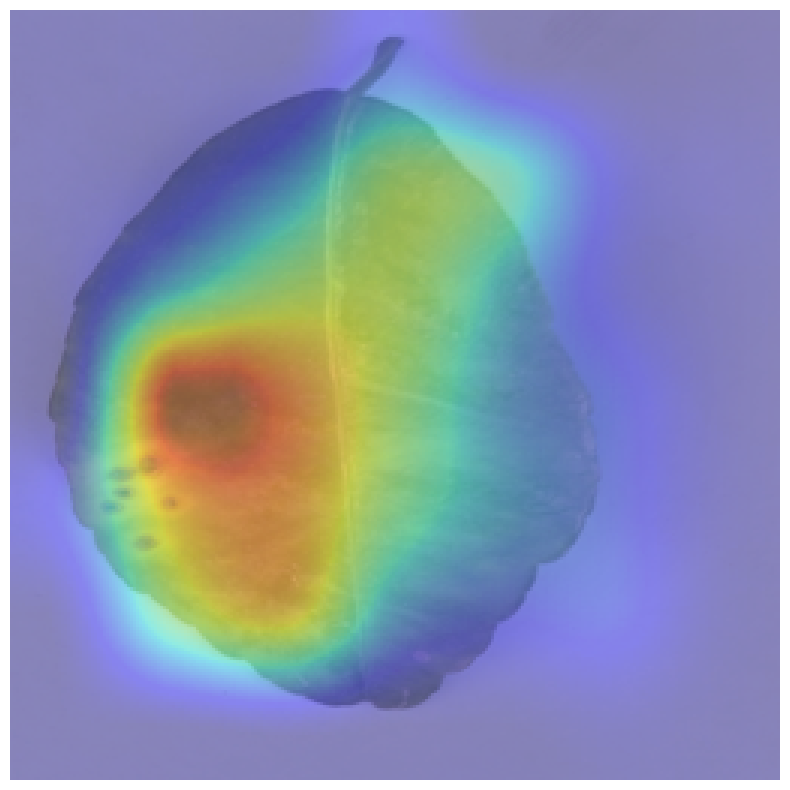} &
        \includegraphics[width=0.14\textwidth]{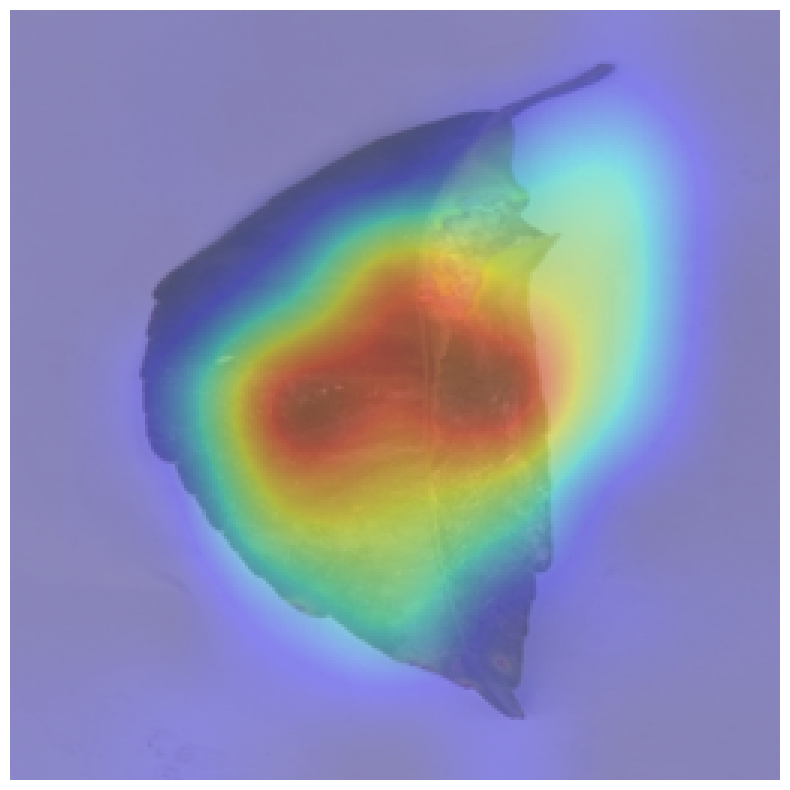} \\ 

        \rotatebox{90}{\scriptsize \text{~~~ConMamba}} &  
        \includegraphics[width=0.14\textwidth]{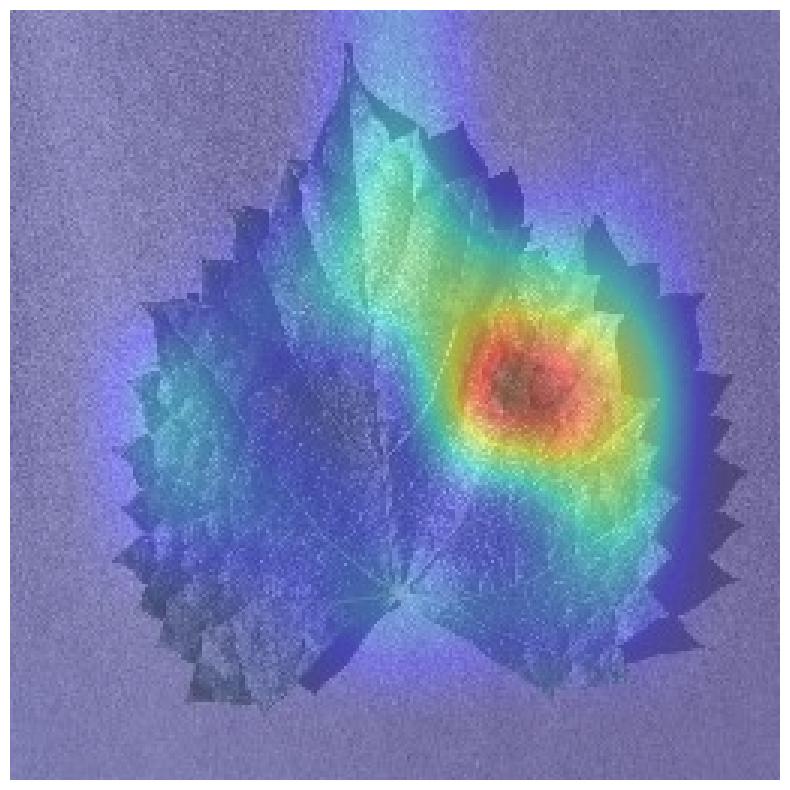} &
        \includegraphics[width=0.14\textwidth]{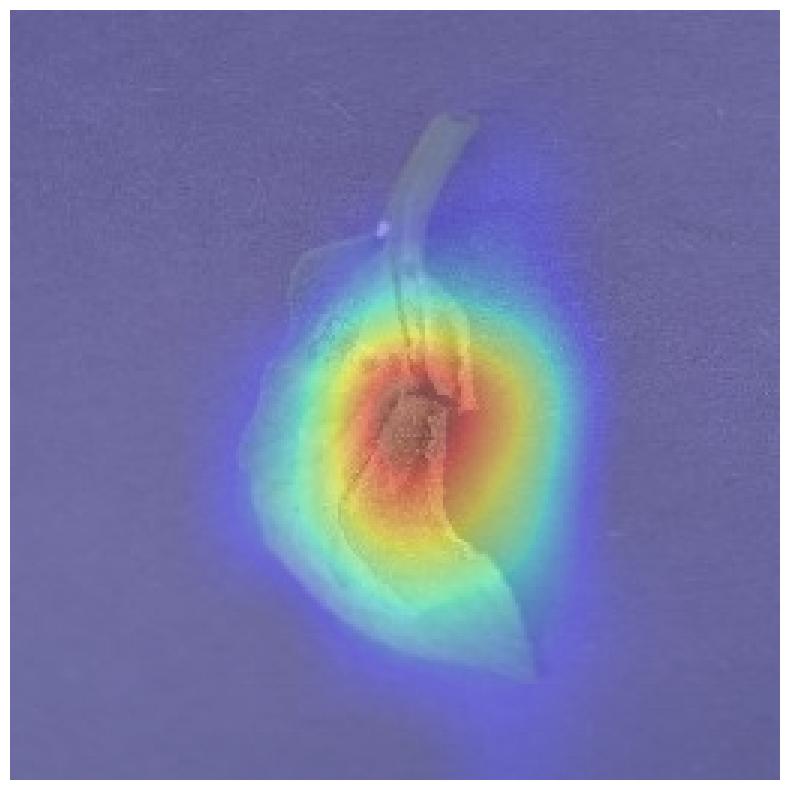} & 
        \includegraphics[width=0.14\textwidth]{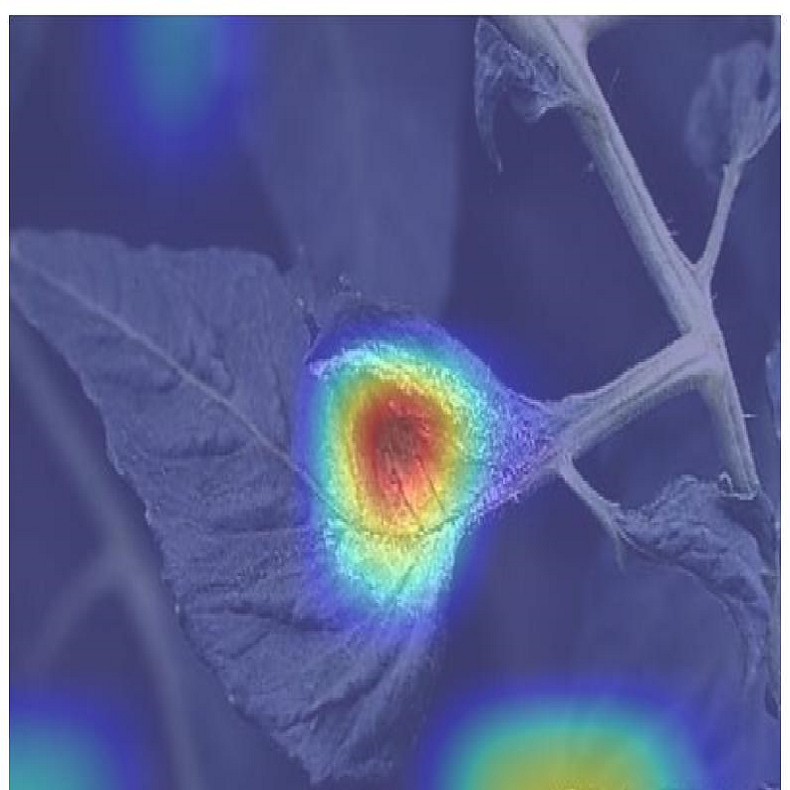} &
        \includegraphics[width=0.14\textwidth]{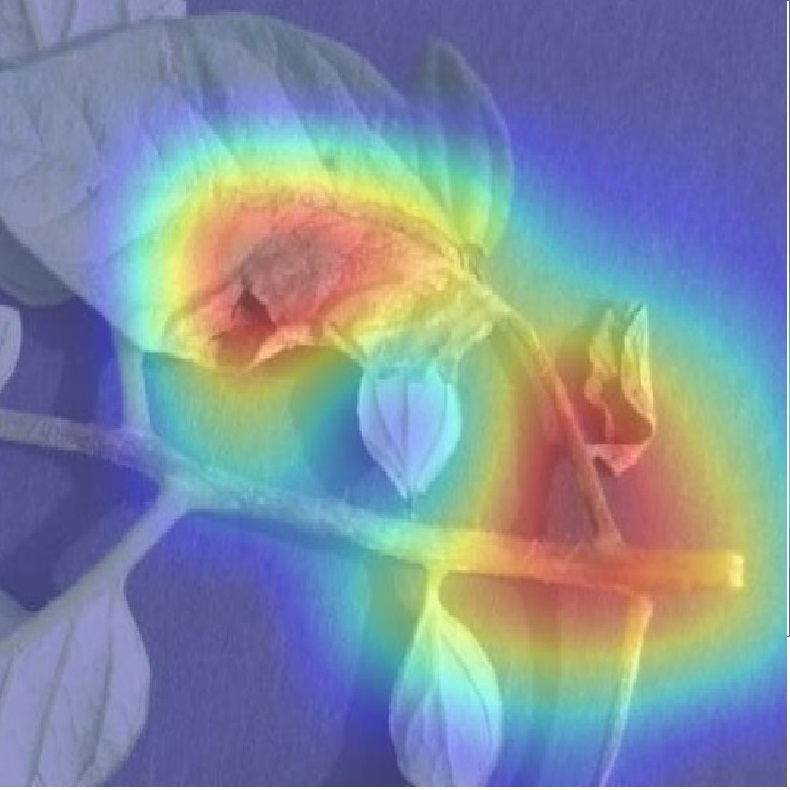} & 
        \includegraphics[width=0.14\textwidth]{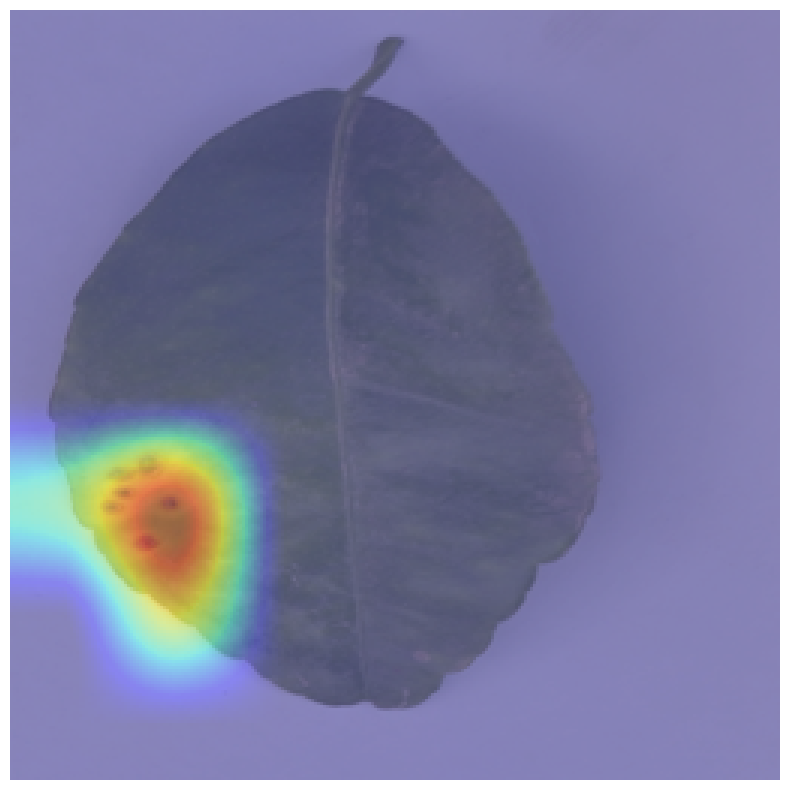} &
        \includegraphics[width=0.14\textwidth]{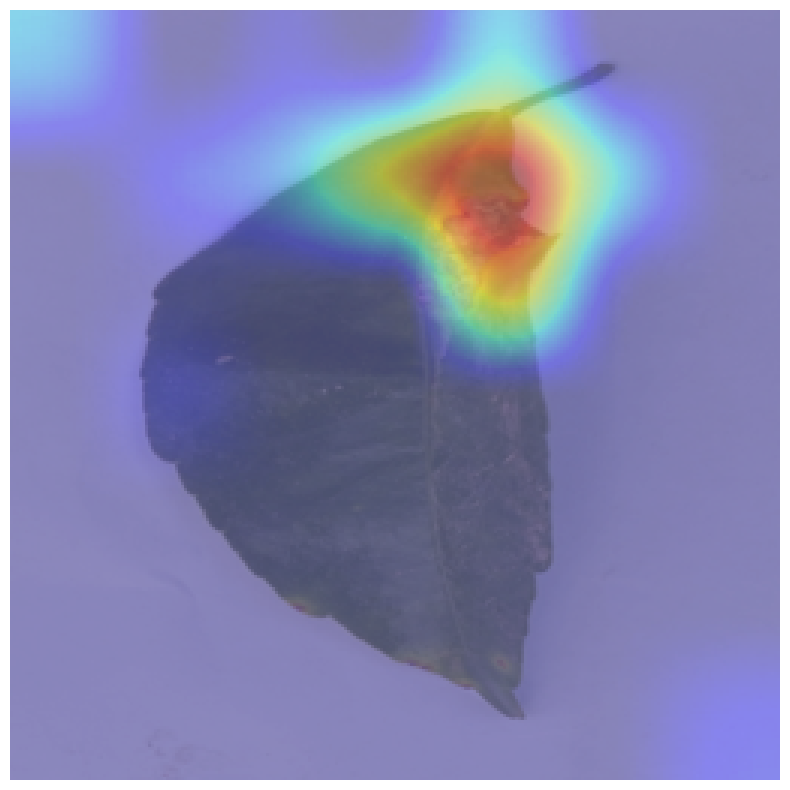} \\ 

    \end{tabular}
    \vspace{0cm} 
    \begin{tabular}{m{5cm} m{5cm} m{5cm}} 
        \centering {\scriptsize \text{~~~~~~~~~~~~~~PlantVillage}} & \centering {\scriptsize \text{~~~~~~~~~~~~PlantDoc}} & \centering {\scriptsize \text{~~~~~~~~~~Citrus}}
    \end{tabular}
    \caption{Visualization of activation maps for randomly chosen images from the PlantVillage, PlantDoc, and Citrus datasets. The first row displays the original samples, followed by the second-best method for each dataset (CAE \& CNN \cite{bedi2021plant} in PlantVillage, CNN \cite{chung2024addressing} in PlantDoc, and Clustering \cite{monowar2022self} in Citrus) and the outcomes produced by our proposed ConMamba approach.}
    \label{fig:gradcam}
\end{figure}

In contrast, the PlantDoc dataset introduces greater complexity due to real world image acquisition artifacts. Here, ConMamba maintains a clear performance margin by achieving 94.29\% accuracy. It exceeds MobileNetV3 \cite{nguyen2023effective} (83.00\%) and CNN \cite{chung2024addressing} (87.42\%) by over 11\% and 6\%, respectively. Notably, the precision (93.88\%), recall (93.87\%), and F1-score (93.97\%) of ConMamba remain consistently superior.

On the Citrus dataset, which contains class distributions with higher interclass similarity and subtle disease manifestations, ConMamba again demonstrates strong performance, achieving the highest accuracy (91.38\%) among all methods evaluated. Compared to prior contrastive and clustering based approaches such as SimSiam \cite{chen2021exploring} (87.50\%), BYOL \cite{grill2020bootstrap} (89.00\%), and Clustering \cite{monowar2022self} (89.30\%), ConMamba achieves superior precision (91.74\%), recall (90.51\%), and F1-score (91.66\%). These results affirm its generalization ability under intraclass variability.

Because the recent transformer-based SSL frameworks for PDD did not report results on our three primary evaluation datasets, we further benchmarked ConMamba on their publicly available test datasets (Table~\ref{tab:comparison}). On the Tea Sickness dataset \cite{shashwatwork_tea_leafs}, ConMamba achieves 95.27\% accuracy, 94.89\% precision, 94.75\% recall, and 94.82\% F1-score, outperforming the transformer-based IEM-ViT \cite{zhang2023information} by 1.49, 1.22, 0.95, and 1.18 percentage points, respectively. Similarly, on the Apple Foliar dataset \cite{thapa2020plant}, ConMamba attains 98.05\% accuracy, 97.55\% precision, 97.62\% recall, and 97.57\% F1-score, surpassing the transformer-driven Mix-ViT \cite{yu2023mix} by 0.69, 0.64, 0.48, and 0.48 percentage points across the respective metrics. These performances across distinct datasets clearly demonstrate that ConMamba outperforms transformer-based self-supervised models, delivering superior class discrimination and generalization capability.

\setlength{\subfigcapskip}{3mm}
\begin{figure}[!t]
    \centering
    \begin{tabular}{cc}
        \subfigure[Second-best method \cite{bedi2021plant}]{\includegraphics[width=0.4\textwidth]{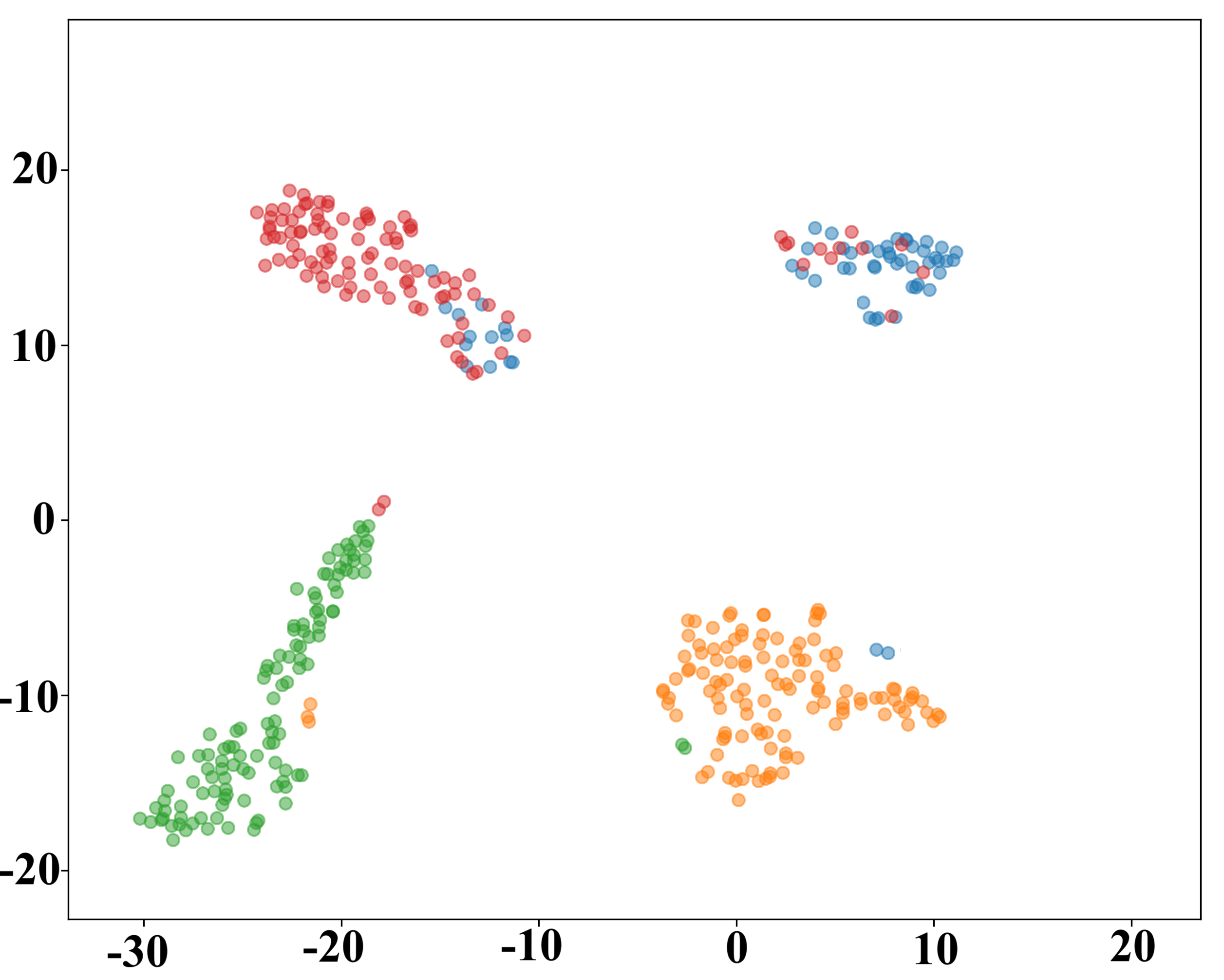}} &
        \subfigure[ConMamba using intra-class contrast]{\includegraphics[width=0.4\textwidth]{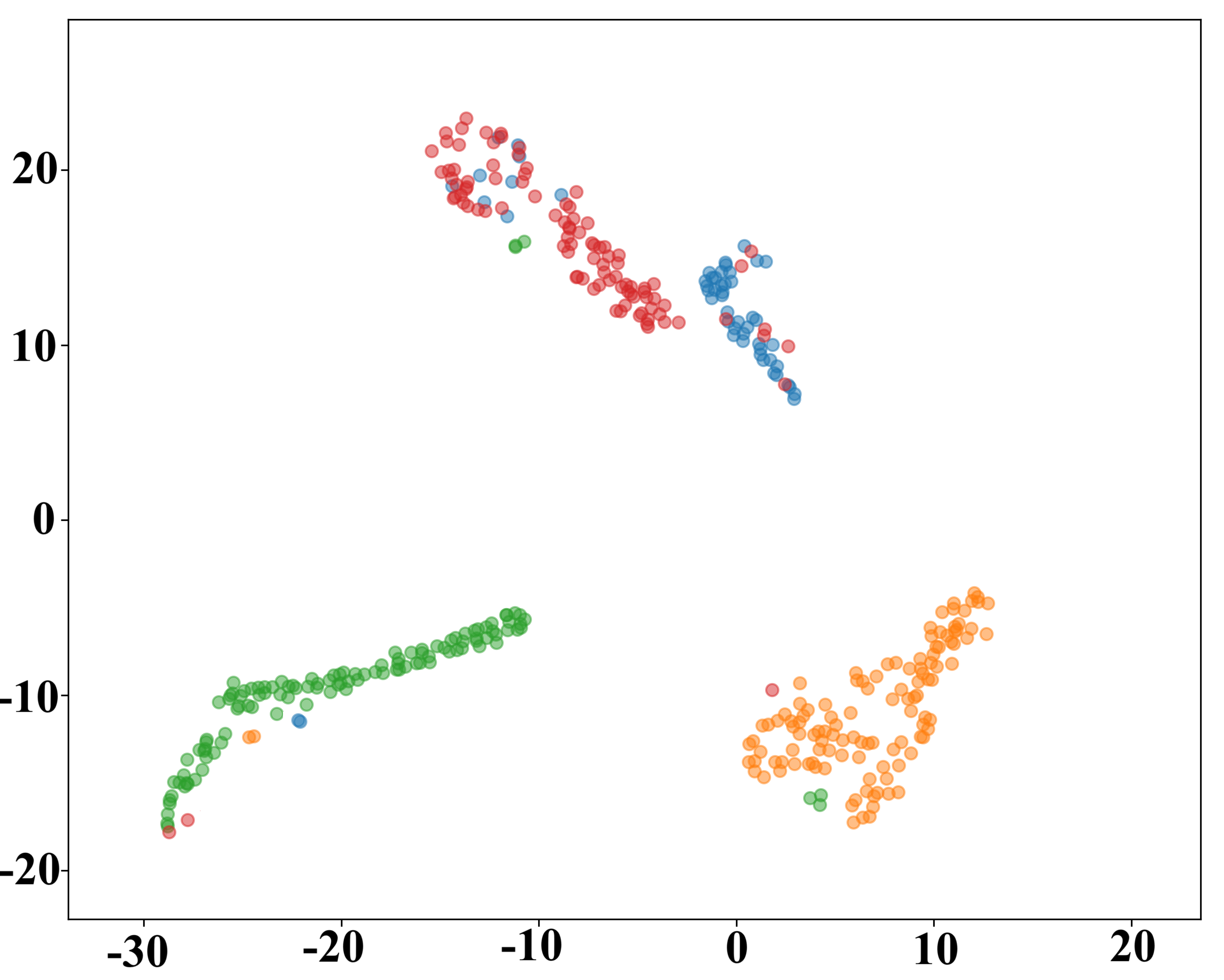}} \\
        \subfigure[ConMamba using inter-class contrast]{\includegraphics[width=0.4\textwidth]{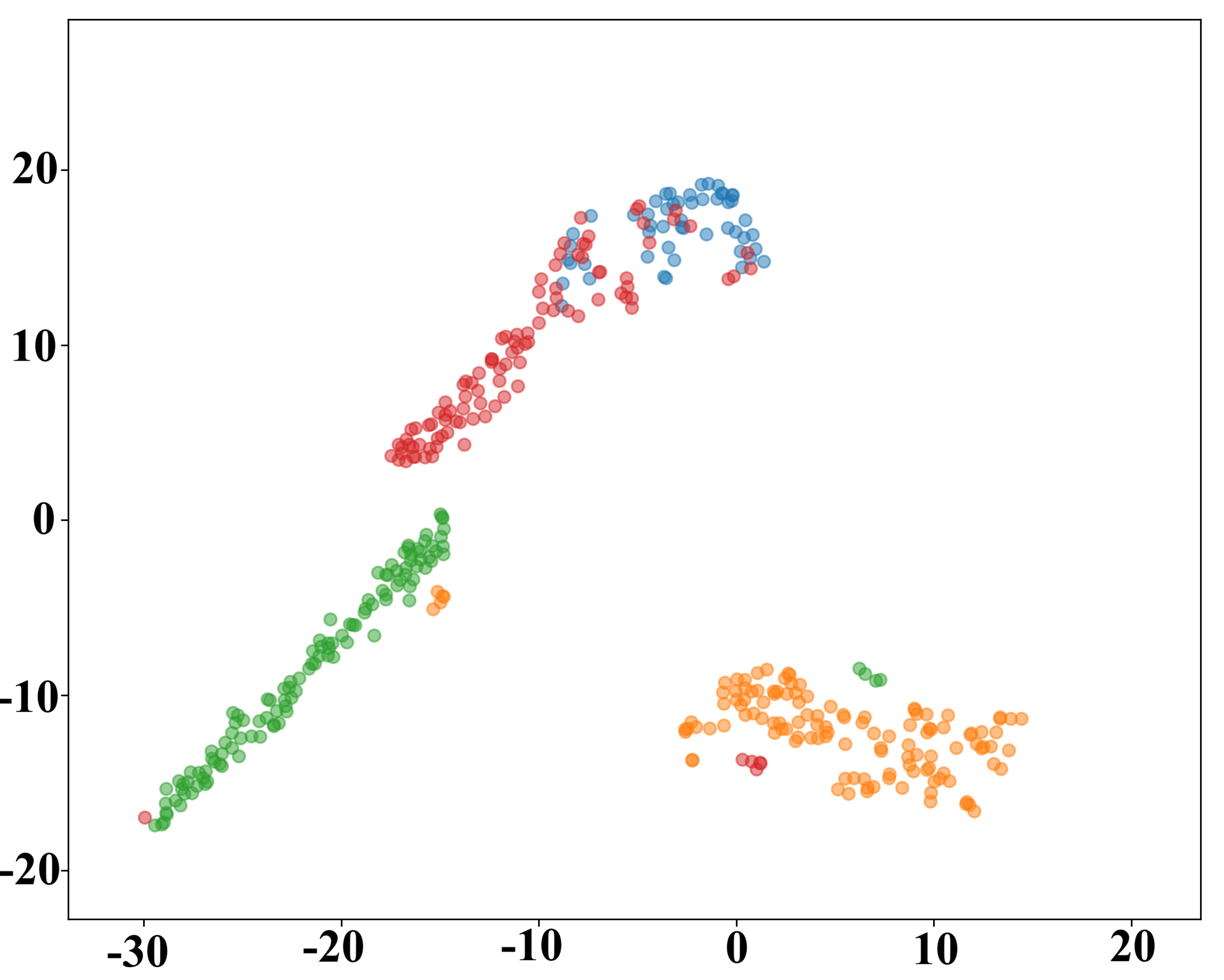}} & 

        \subfigure[ConMamba using dual-level contrastive loss]{\includegraphics[width=0.4\textwidth]{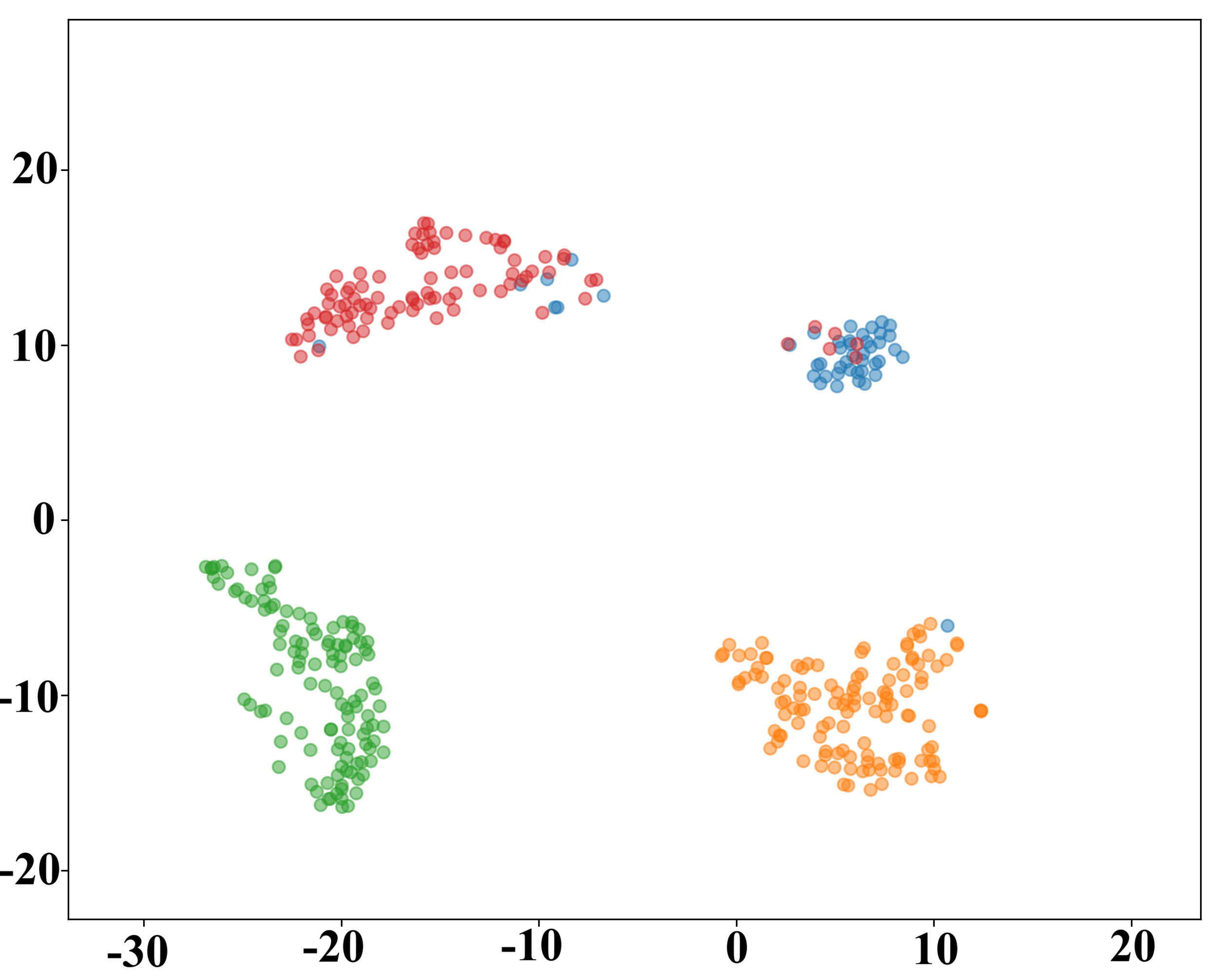}} \\

    \end{tabular}
    \caption{t-SNE visualization of feature representations learned by ConMamba on \textbf{corn plant} images from the \textbf{PlantVillage dataset}. The plots show the embeddings: (a) second-best method \cite{bedi2021plant}, (b) ConMamba using intra-class contrast, (c) ConMamba using inter-class contrast, and (d) ConMamba using dual-level contrastive loss. The figure uses different colors to represent four distinct classes of corn plants. The X and Y axes correspond to the t-SNE Components representing transformed dimensions from the high-dimensional data.}
    \label{fig:tsne_PV}
    \vspace{-9pt}
\end{figure}

These consistent improvements across all datasets are attributed to ConMamba’s rich feature representation capabilities, which effectively capture both fine-grained local features and global long-range dependencies. Besides, its dual-level contrastive learning mechanism aligns features at both local(intra-class) and global(inter-class) levels. Moreover, dynamic uncertainty-based loss balancing allows adaptive learning focus, enhancing feature discriminability. In contrast, existing approaches often rely on static loss designs and traditional convolutional backbones, which lack the capacity to model hierarchical spatial relationships and adapt dynamically during training. ConMamba overcomes these limitations, offering a unified and robust framework for self-supervised PDD.

\subsubsection{Qualitative Performance}
To qualitatively assess the discriminative capacity of ConMamba, we generate Class Activation Maps(CAMs)~\cite{yu2021maskcov} based on the model's predictions during the testing stage. These visualizations highlight the regions most influential in the model's decision-making, and show how well it localizes disease symptoms.
This visualization provides qualitative evidence of the model's ability to accurately localize disease-affected areas and differentiate between healthy and diseased regions.  
Figure \ref{fig:gradcam} presents a side-by-side comparison between ConMamba and the second-best performing methods across three datasets. For each dataset, two representative images are randomly selected. As shown in the figure, ConMamba consistently attends to the most relevant symptomatic regions, while the baseline model often exhibit over-extended attention regions or focuses on irrelevant areas, leading to misinterpretation and reduced localization precision.  These results confirm that ConMamba's long-range feature modeling and dual-level contrastive learning contribute to more targeted and context-aware spatial attention, which is critical for reliable and fine-grained plant disease classification.

\begin{figure}[!t]
    \centering
    \begin{tabular}{cc}
        \subfigure[Second-best method \cite{monowar2022self}]{\includegraphics[width=0.4\textwidth]{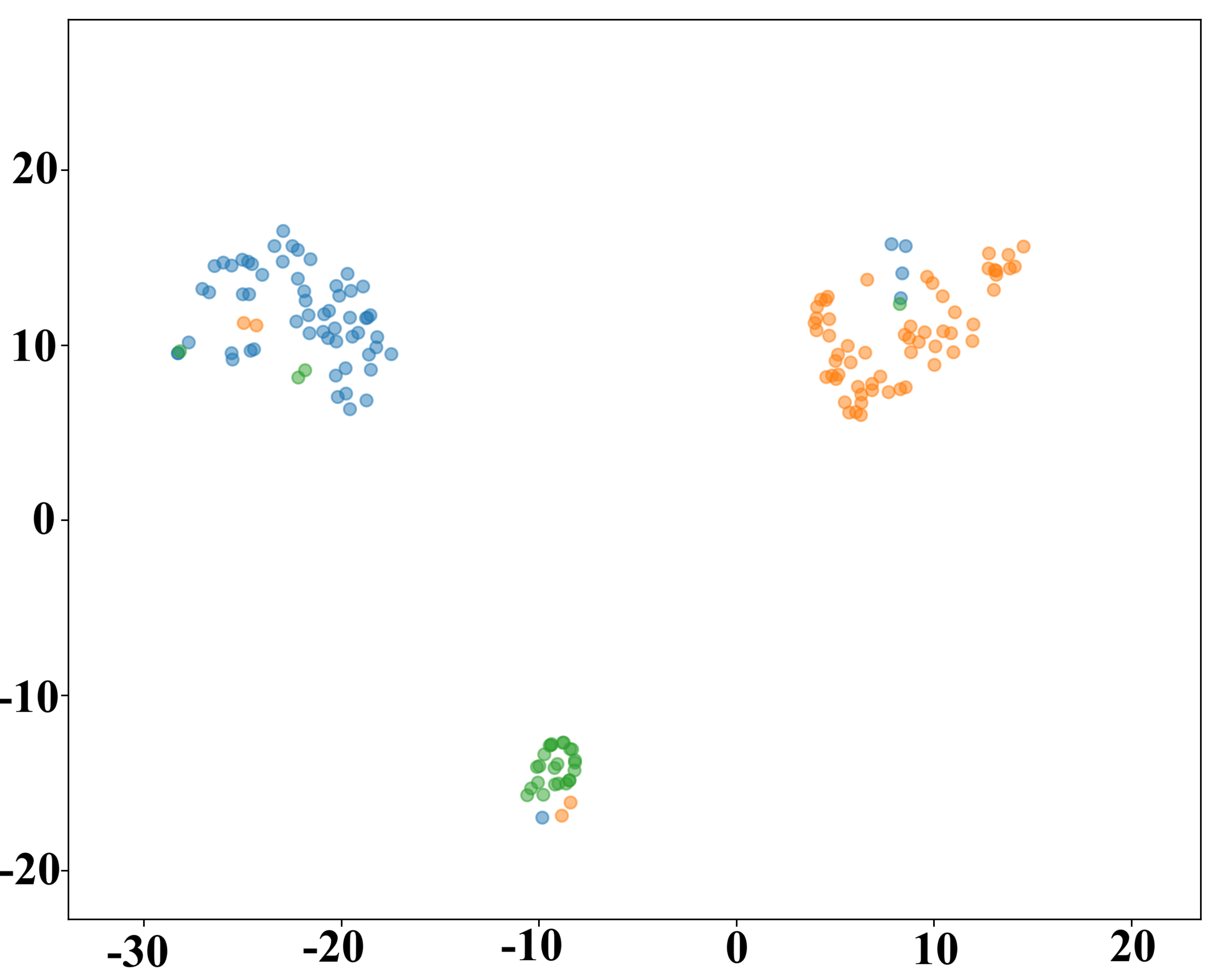}} &
        \subfigure[ConMamba using intra-class contrast]{\includegraphics[width=0.4\textwidth]{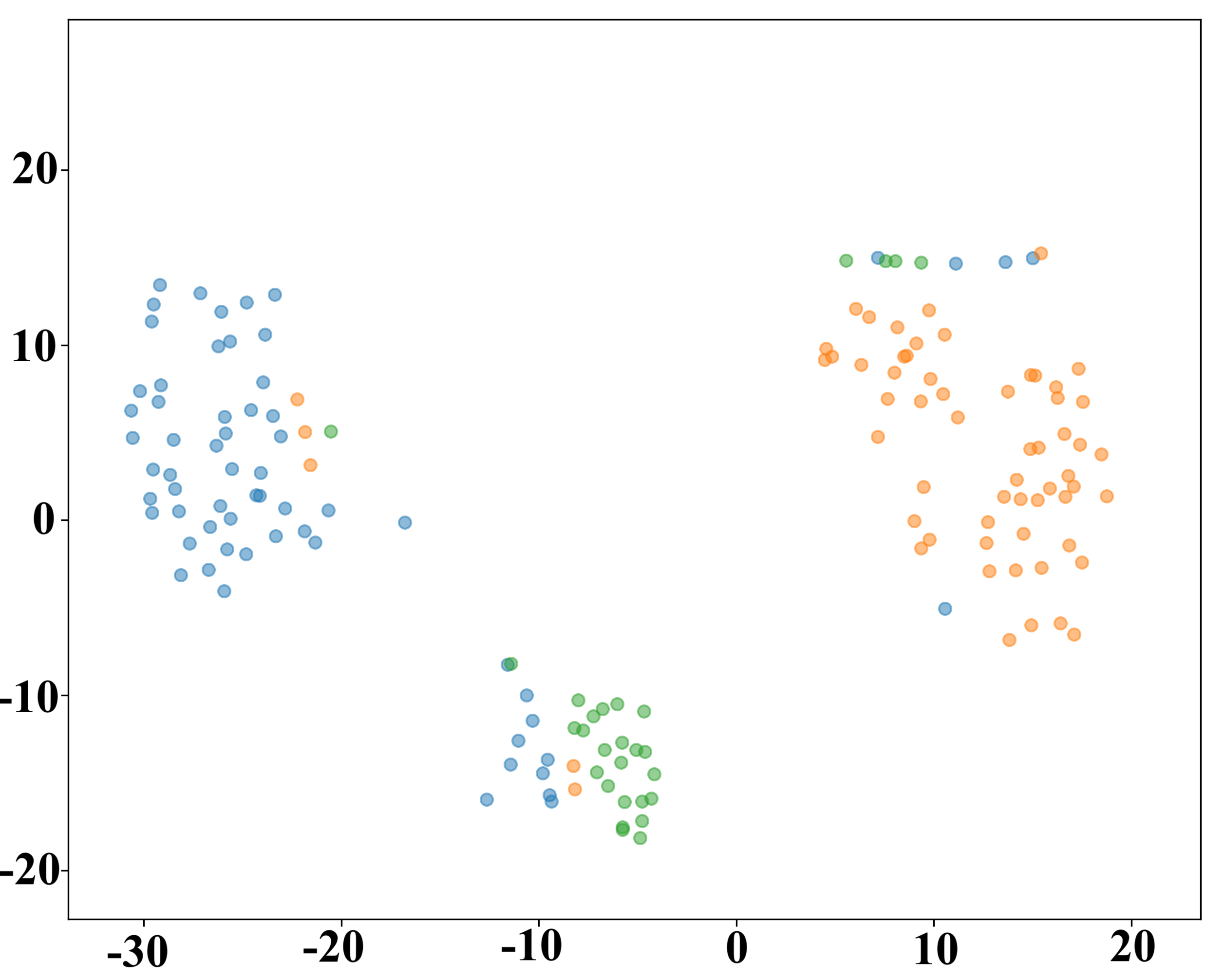}} \\
        \subfigure[ConMamba using inter-class contrast]{\includegraphics[width=0.4\textwidth]{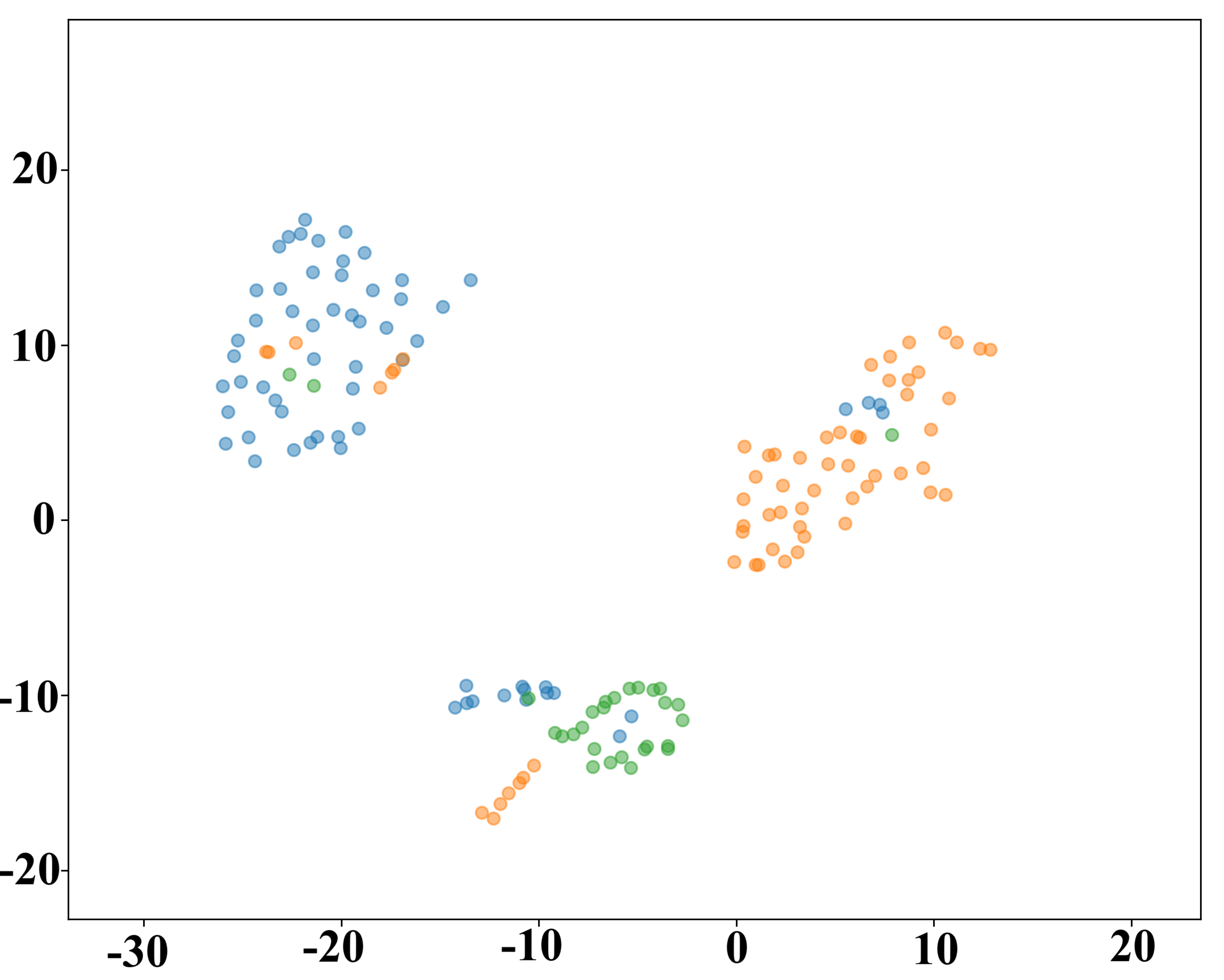}} & 

        \subfigure[ConMamba using dual-level contrastive loss]{\includegraphics[width=0.4\textwidth]{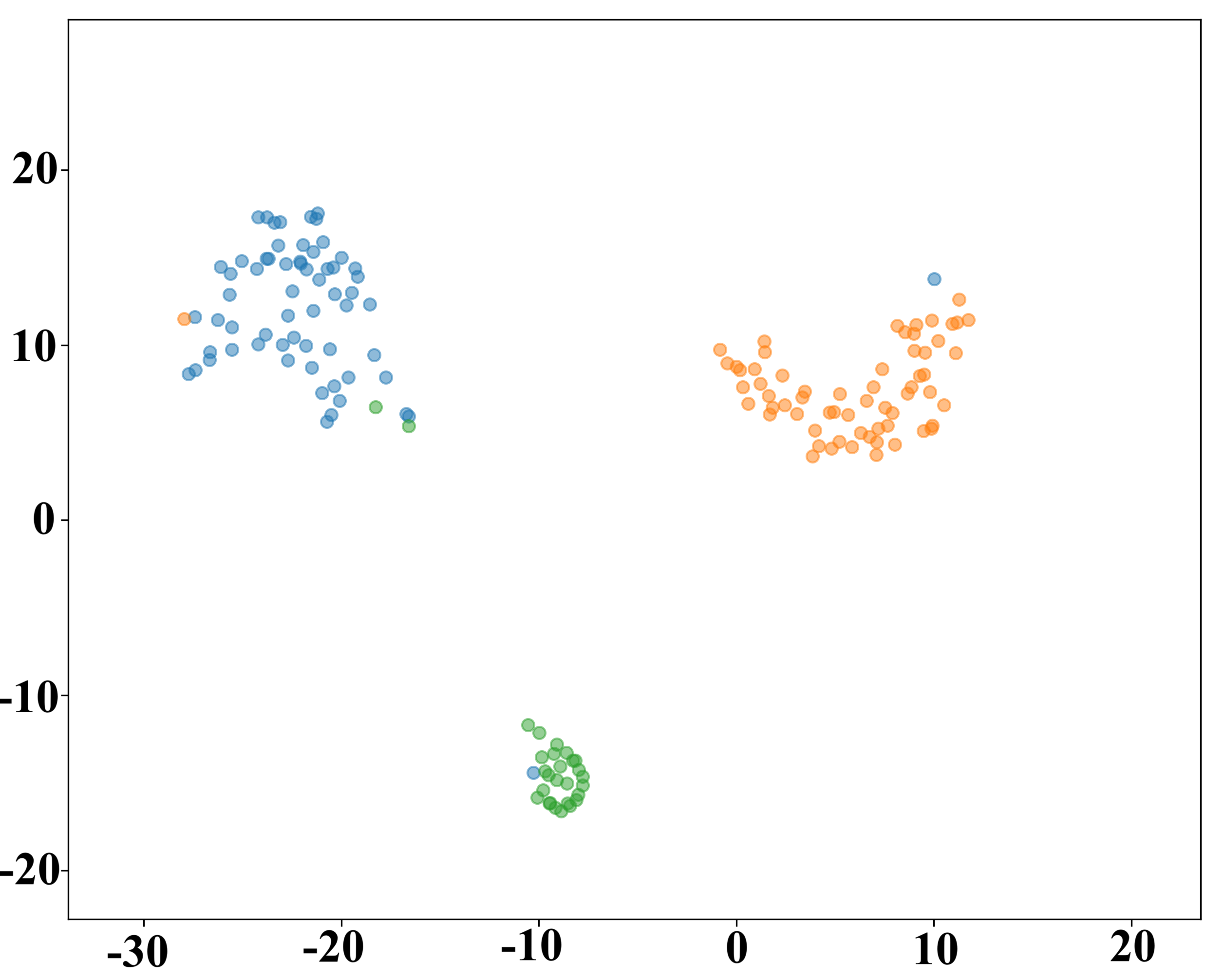}} \\
 
    \end{tabular}
    \caption{t-SNE visualization of feature representations learned by ConMamba on \textbf{corn plant} images from the \textbf{PlantDoc} dataset. The plots show the embeddings: (a) second-best method \cite{monowar2022self}, (b) ConMamba using intra-class contrast, (c) ConMamba using inter-class contrast, and (d) ConMamba using dual-level contrastive loss. The figure uses different colors to represent three distinct classes. The X and Y axes correspond to the t-SNE Components representing transformed dimensions from the high-dimensional data.}
    \label{fig:tsne_PD}
    \vspace{-9pt}
\end{figure}

\subsection{Ablation Studies}
\subsubsection{Loss Functions} \label{sec:loss_function}
In this ablation study, we perform a discriminative class analysis using t-SNE \cite{maaten2008visualizing} to evaluate the effectiveness of the proposed ConMamba model in the test datasets. 
t-SNE is a non-linear dimensionality reduction technique that visualizes high-dimensional data in two dimensions while preserving local similarities, helping to reveal classification patterns and clusters. 

\begin{figure}[!t]
    \centering
    \begin{tabular}{cc}
        \subfigure[Second-best method \cite{monowar2022self}]{\includegraphics[width=0.4\textwidth]{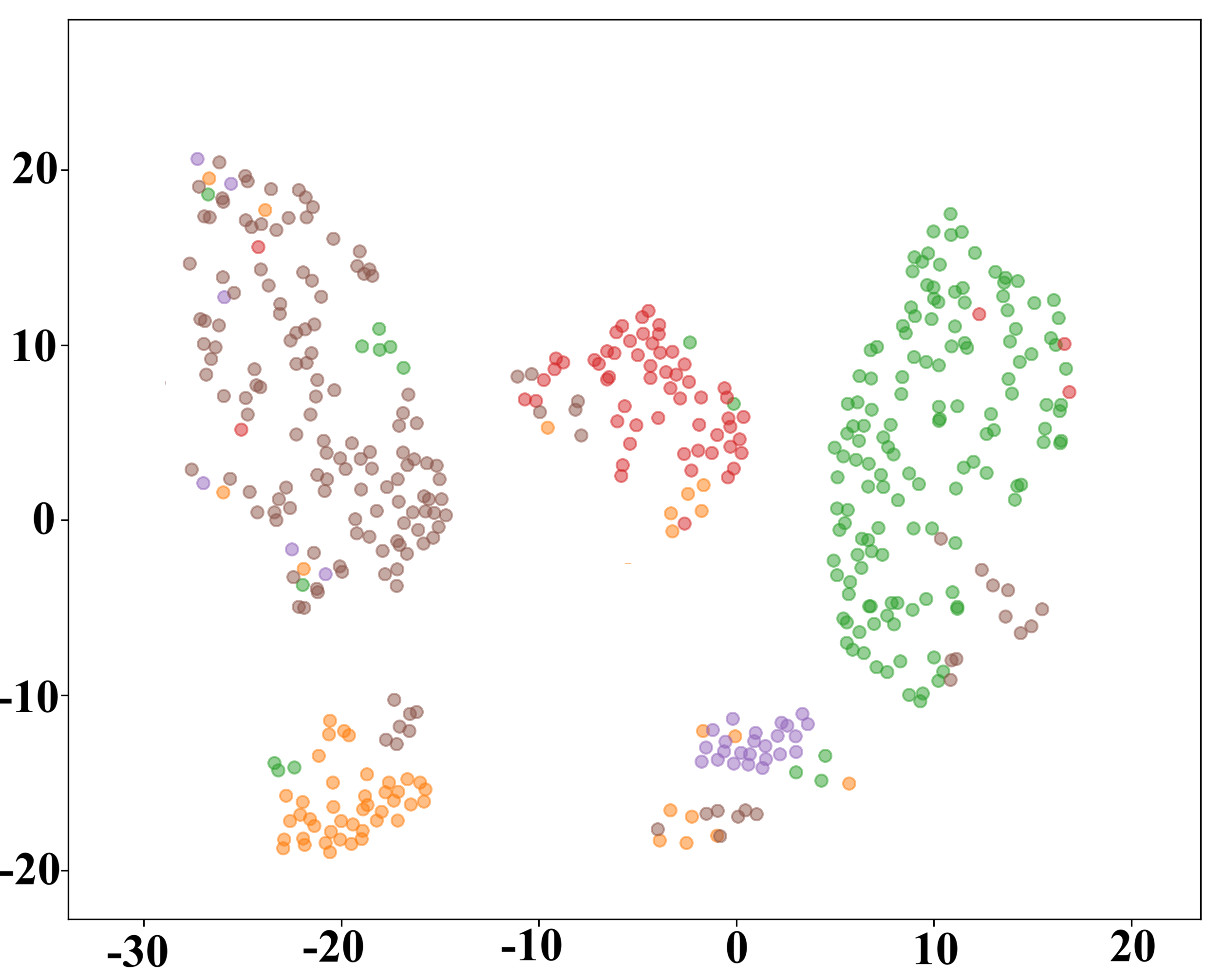}} &        \subfigure[ConMamba using intra-class contrast]{\includegraphics[width=0.4\textwidth]{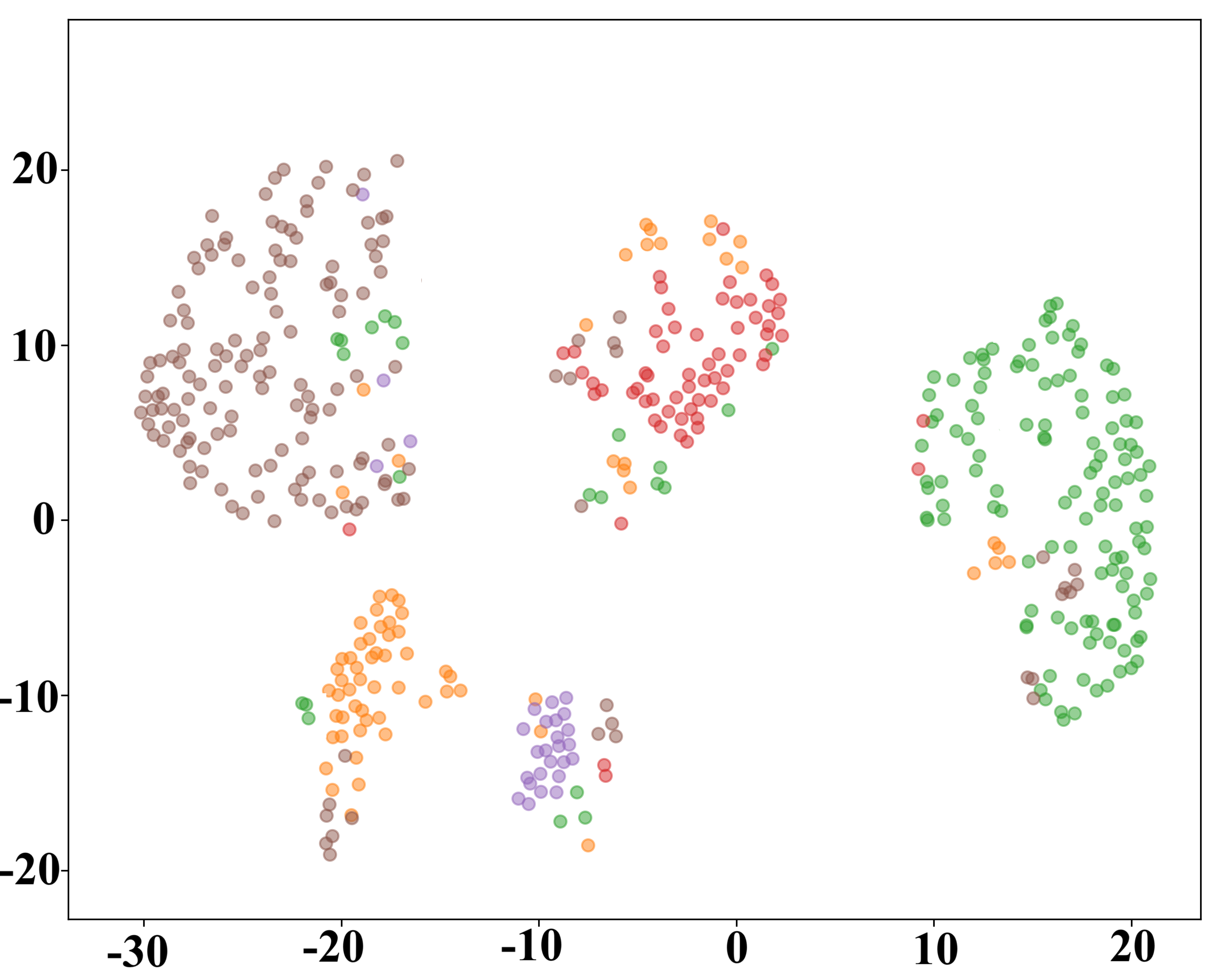}} \\
        \subfigure[ConMamba using inter-class contrast]{\includegraphics[width=0.4\textwidth]{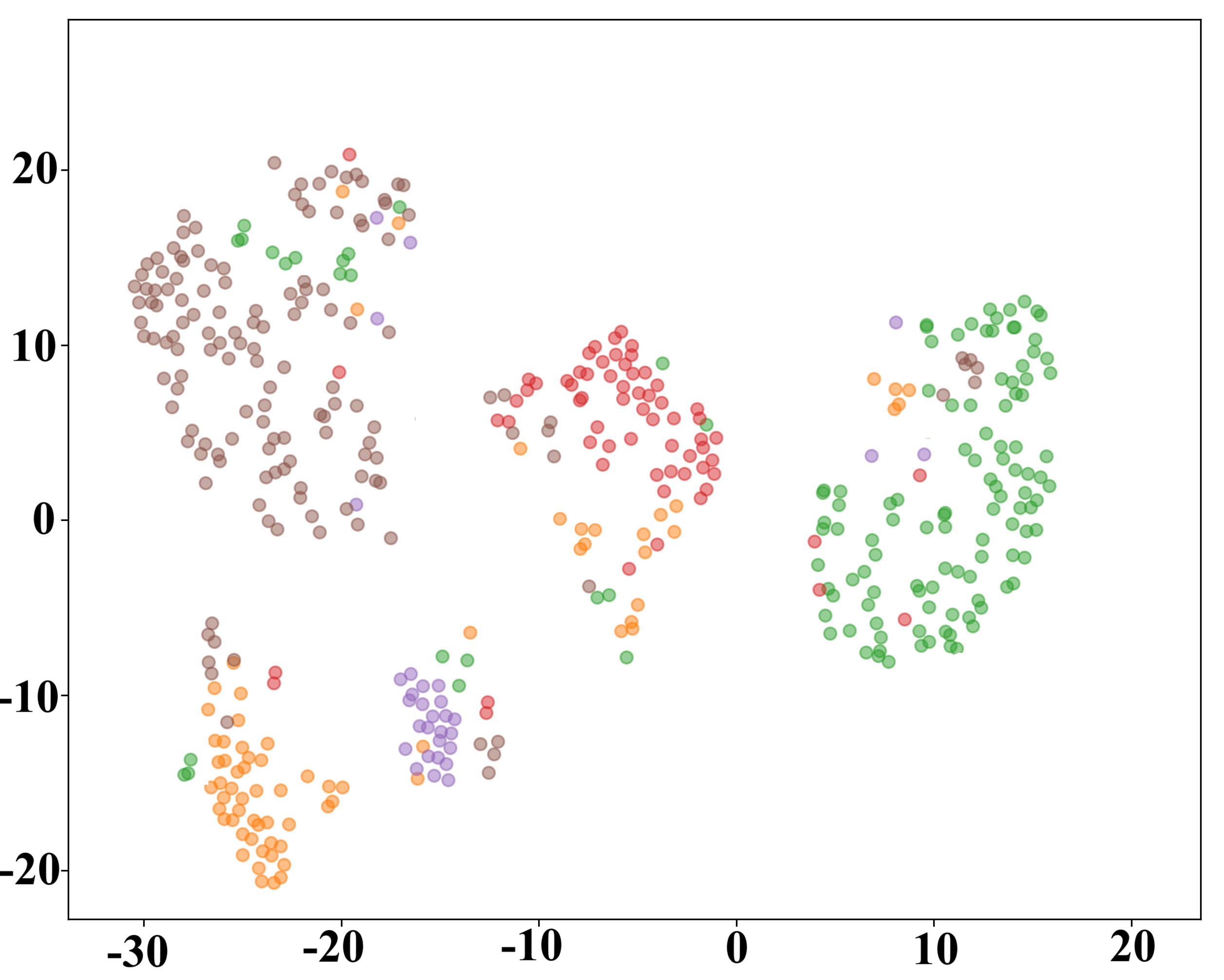}} & 

        \subfigure[ConMamba using dual-level contrastive loss]{\includegraphics[width=0.4\textwidth]{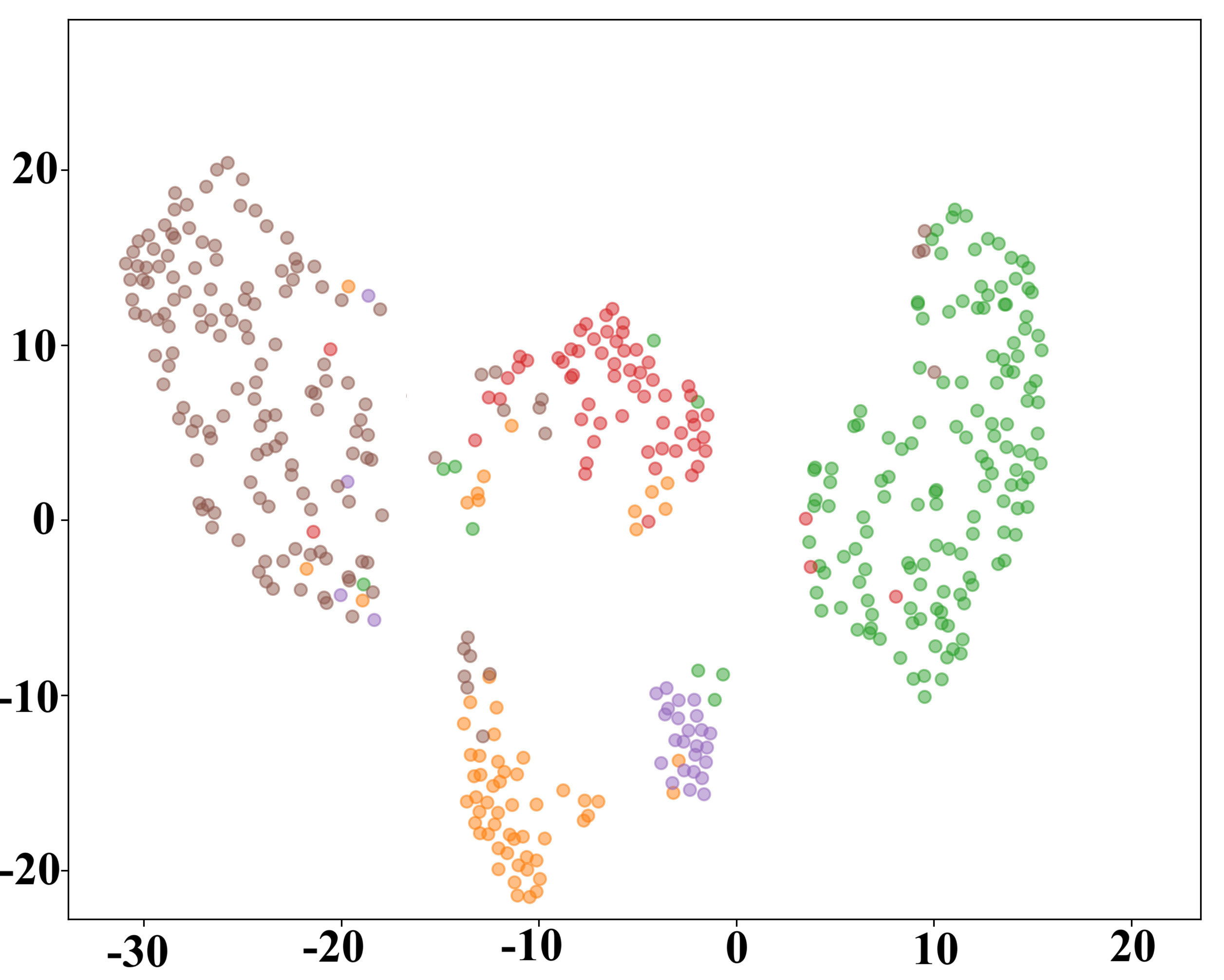}} \\
    \end{tabular}
    \caption{t-SNE visualization of feature representations learned by ConMamba on the \textbf{Citrus} dataset. The plots show the embeddings: (a) the second-best method \cite{monowar2022self}, (b) ConMamba using intra-class contrast, (c) ConMamba using inter-class contrast, and (d) ConMamba using dual-level contrastive loss. The figure uses different colors to represent five distinct classes. The X and Y axes correspond to the t-SNE Components representing transformed dimensions from the high-dimensional data.}
    \label{fig:tsne_citrus}
    \vspace{-9pt}
\end{figure}

Since the PlantVillage and PlantDoc datasets contain numerous classes (38 and 28, respectively), we selected only the corn plant classes from each dataset to achieve clearer visualisations using t-SNE. Figures \ref{fig:tsne_PV}–\ref{fig:tsne_citrus} illustrate the discriminative capabilities of the ConMamba model employing individual loss functions (intra-class contrast and inter-class contrast) compared to the proposed dual-level contrastive loss, alongside the second-best performing models for each dataset.

In Figure \ref{fig:tsne_PV} (PlantVillage dataset), our proposed dual-level contrastive loss (d) demonstrates the best feature separation, forming well-defined clusters that indicate superior feature learning.
Compared to intra-class contrast (b) and inter-class contrast (c), the dual-level contrastive loss produces more clearer cluster boundaries with minimal overlap. In contrast, the second-best method \cite{bedi2021plant} (a) exhibits noticeable cluster overlap, indicating weaker feature extraction. Moving to Figure \ref{fig:tsne_PD} (PlantDoc dataset), the overall separability of clusters is less distinct than in the PlantVillage dataset, suggesting that PlantDoc is a more challenging dataset. However, our dual-level contrastive loss (d) still outperforms the individual losses and second-best method \cite{monowar2022self} (a), maintaining clear separation and reducing misclassified points. Finally, in Figure \ref{fig:tsne_citrus} (Citrus dataset), the dual-level contrastive loss (d) again provides the best cluster separation, with clear distinctions between the classes. While the individual losses and second-best method \cite{monowar2022self} (a) exhibit significant cluster overlaps, reinforcing the effectiveness of the dual-level contrastive learning approach.

The improvement trend across all datasets consistently highlights that ConMamba, with the proposed dual-level contrastive loss, substantially improves class discrimination, resulting in superior separation among various plant disease categories relative to alternative approaches.

\begin{table}[!t]
\scriptsize
\centering
\caption{Comparative evaluation of ConMamba and the second-best performing models of each test dataset under random and balanced sampling strategies. The metrics include Accuracy, Macro F1, and Minority F1.}
    \vspace{-3pt}
\label{tab:Class Imbalance}
\begin{tblr}{
  cells = {c},
  cell{2}{1} = {r=4}{},
  cell{2}{2} = {r=2}{},
  cell{4}{2} = {r=2}{},
  cell{6}{1} = {r=4}{},
  cell{6}{2} = {r=2}{},
  cell{8}{2} = {r=2}{},
  cell{10}{1} = {r=4}{},
  cell{10}{2} = {r=2}{},
  cell{12}{2} = {r=2}{},
  hline{1-2,14} = {-}{},
  hline{4,6,8,10,12} = {3-6}{},
}
\textbf{Dataset}      & \textbf{Model}    & \textbf{Sampling} & \textbf{Accuracy} & {\textbf{Macro F1}} & {\textbf{Minority F1}} \\
PlantVillage \cite{al2024plant} & {CAE \& CNN \cite{bedi2021plant}      ~} & Random            & 98.38             & 96.85                         & 95.12                            \\
                      &                   & Balanced          & 98.52             & 97.12                         & 95.84                            \\
                      & ConMamba          & Random            & 98.62             & 97.38                         & 96.12                            \\
                      &                   & Balanced          & \textbf{98.71}    & \textbf{97.91}                & \textbf{97.23}                   \\
PlantDoc \cite{Singh2020PlantDoc}     & CNN \cite{chung2024addressing}               & Random            & 87.42             & 84.70                         & 77.85                            \\
                      &                   & Balanced          & 88.21             & 86.34                         & 80.42                            \\
                      & ConMamba          & Random            & 94.29             & 91.23                         & 89.15                            \\
                      &                   & Balanced          & \textbf{94.56}    & \textbf{92.84}                & \textbf{91.71}                   \\
Citrus \cite{rauf2019citrus}       & Clustering \cite{monowar2022self} ~      & Random            & 89.30             & 86.11                         & 81.03                            \\
                      &                   & Balanced          & 89.75             & 87.05                         & 83.29                            \\
                      & ConMamba          & Random            & 91.38             & 91.66                         & 89.42                            \\
                      &                   & Balanced          & \textbf{91.80}    & \textbf{92.17}                & \textbf{91.01}                   
\end{tblr}
\end{table}

\subsubsection{Handling Class Imbalance}
Real-world plant disease datasets, such as PlantDoc and Citrus, often exhibit significant class imbalance, where a few dominant classes account for the majority of samples while others remain underrepresented. To evaluate the robustness of ConMamba in such scenarios, we investigate both model-level mechanisms and data-level strategies for mitigating imbalance, supported by imbalance-aware evaluation metrics.

\noindent \textbf{ Model-Level Imbalance Mitigation:}
ConMamba addresses class imbalance through a combination of complementary architectural strategies designed to improve feature learning across both majority and minority classes. First, the intra-class contrastive learning component encourages alignment of augmented views from the same instance, which helps the model learn consistent and robust representations even for classes with limited training samples. In parallel, inter-class contrastive learning promotes greater separation between distinct class representations, reducing ambiguity and improving discriminability, particularly between minority and majority categories. Additionally, ConMamba incorporates an uncertainty-guided dynamic loss weighting mechanism, which adaptively emphasizes samples with high uncertainty. These samples often correspond to harder or less frequent classes, and prioritizing them during training reduces the risk of overfitting to dominant classes. These mechanisms enable ConMamba to learn more balanced and generalizable feature representations without requiring explicit class reweighting or resampling.

\noindent \textbf{Data-Level Sampling Strategy:}
To complement our model-level strategies, we perform experiments comparing random sampling (reflecting natural class imbalance) with balanced sampling (equalizing class presence in each mini-batch) during training. While ConMamba performs well under both settings, balanced sampling provides more uniform exposure to underrepresented classes. As shown in Table \ref{tab:Class Imbalance}, ConMamba achieves higher macro-averaged F1-scores and improved accuracy under balanced sampling, with minimal impact on majority class performance. These results confirm that the model can effectively handle real-world imbalance and benefit from additional support at the data level. To validate the advantage of ConMamba over existing methods, we compare it against the second-best baseline of each test dataset. under both sampling conditions. The results show that ConMamba outperforms the baseline under both strategies, demonstrating stronger robustness and better generalization to minority classes.

\begin{table}[!t]
\centering
\caption{Performance comparison of Vision Mamba and traditional encoder using dual-level contrastive loss on the \textbf{PlantVillage} dataset. Evaluation metrics include Accuracy (\%), Recall (\%), Precision (\%), and F1-score (\%).}
    \vspace{-3pt}
\label{tab:ablation plantvillage}
\resizebox{0.6\linewidth}{!}{
\begin{tblr}{
  cells = {c},
  hline{1-2,10} = {-}{},
}
\textbf{Method}  & \textbf{Accuracy} & \textbf{Recall} & \textbf{Precision} & \textbf{F1-score} \\
AlexNet \cite{krizhevsky2017imagenet}         & 81.55             & 81.89           & 81.33              & 80.4              \\
MobileNetV3Large \cite{howard2019searching}& 80.75             & 76.03           & 76.67              & 81.34             \\
EfficientNetB0 \cite{tan2019efficientnet}   & 91.66             & 88.15           & 90.31              & 90.68             \\
VGG16 \cite{simonyan2014very}           & 83.55             & 81.89           & 82.33              & 83.04             \\
ResNet50 \cite{he2016deep}        & 93.35             & 92.5            & 91.23              & 90.65             \\

ViT \cite{dosovitskiy2020image}             & 94.95             & 94.67           & 95.55              & 95.87             \\
Swin-T \cite{liu2021swin}             & 96.17             & 95.88           & 95.72              & 96.01             \\
\textbf{VM \cite{zhu2024vision}}              & \textbf{98.62}    & \textbf{97.59}           & \textbf{96.74}              & \textbf{97.38 }            
\end{tblr}}
\caption*{\textbf{Note:} ViT = Vision Transformer; Swin-T = Swin Transformer; VM = Vision Mamba.}
\vspace{-12pt}
\end{table}

\noindent \textbf{Feature Distribution Visualization:}
As shown in Section \ref{sec:loss_function}, where t-SNE visualizations (Figures \ref{fig:tsne_PV}-\ref{fig:tsne_citrus}) of learned embeddings illustrate the representational quality of different configurations. In particular, ConMamba with dual-level contrastive learning forms more compact and well-separated class clusters, including those for underrepresented classes. This supports the claim that our model architecture helps preserve meaningful class structure even under imbalance.

\begin{table}[!t]
\centering
\caption{Performance comparison of Vision Mamba and traditional encoder using dual-level contrastive loss on the \textbf{PlantDoc} dataset. Evaluation metrics include Accuracy (\%), Recall (\%), Precision (\%), and F1-score (\%).}
    \vspace{-3pt}
\label{tab:ablation plantdoc}
\resizebox{0.6\linewidth}{!}{
\begin{tblr}{
  cells = {c},
  hline{1-2,10} = {-}{},
}
\textbf{Method}  & \textbf{Accuracy} & \textbf{Recall} & \textbf{Precision} & \textbf{F1-score} \\
AlexNet \cite{krizhevsky2017imagenet}         & 85.27             & 81.75           & 81.82              & 82.24             \\
MobileNetV3Large \cite{howard2019searching} & 87.75             & 86.03           & 86.67              & 86.34             \\
EfficientNetB0 \cite{howard2019searching}  & 88.69             & 87.47           & 85.62              & 86.97             \\
VGG16 \cite{simonyan2014very}           & 87.55             & 86.89           & 87.33              & 87.04             \\
ResNet50 \cite{he2016deep}        & 89.8              & 88.67           & 88.06              & 89.18             \\
ViT \cite{dosovitskiy2020image}             & 91.78             & 90.54           & 91.63              & 91.52             \\
Swin-T \cite{liu2021swin}             & 92.64             & 91.72           & 91.95              & 92.33             \\
\textbf{VM \cite{zhu2024vision}}              & \textbf{94.29}    & \textbf{93.87}  & \textbf{93.88}     & \textbf{93.97}    
\end{tblr}}
\caption*{\textbf{Note:} ViT = Vision Transformer; Swin-T = Swin Transformer; VM = Vision Mamba.}
\vspace{-12pt}
\end{table}

\subsubsection{Encoders with Dual-level Contrastive Learning}

This ablation study presented in Tables \ref{tab:ablation plantvillage}–\ref{tab:ablation citrus} illustrates the effectiveness of the proposed VME with dual-level contrastive loss compared to traditional encoders, such as EfficientNetB0 \cite{howard2019searching}, VGG16 \cite{simonyan2014very}, ResNet50 \cite{he2016deep}, ViT \cite{dosovitskiy2020image}, and Swin Transformer (Swin-T) \cite{liu2021swin} across the evaluated datasets. 
VME consistently achieved superior performance, demonstrating accuracy rates of 98.62\%, 94.29\%, and 91.38\%, along with corresponding F1-scores of 97.38\%, 93.97\%, and 91.66\% on the PlantVillage (Table \ref{tab:ablation plantvillage}), PlantDoc (Table \ref{tab:ablation plantdoc}), and Citrus (Table \ref{tab:ablation citrus}) datasets, respectively. Comparatively, traditional neural network models such as AlexNet \cite{krizhevsky2017imagenet}, MobileNetV3Large \cite{howard2019searching}, EfficientNetB0 \cite{howard2019searching}, VGG16 \cite{simonyan2014very}, and ResNet50 \cite{he2016deep} exhibited lower performance across all metrics. Among these traditional architectures, ResNet50 \cite{he2016deep} and EfficientNetB0 \cite{howard2019searching} performed better than AlexNet \cite{krizhevsky2017imagenet} and MobileNetV3Large \cite{howard2019searching}, though still significantly behind VM. When compared with transformer architectures, ViT \cite{dosovitskiy2020image} and Swin-T \cite{liu2021swin} showed competitive results. In particular, Swin-T \cite{liu2021swin} improved upon ViT \cite{dosovitskiy2020image} in the PlantVillage dataset (96.17\% accuracy, 96.02\% F1-score) and demonstrated solid generalization on the PlantDoc and Citrus datasets (92.64\% and 90.72\% accuracy, respectively). Nonetheless, VME outperformed both transformers across all benchmarks, maintaining at least a 2-3 \% lead in F1-score and exhibiting superior consistency in recall and precision. These observations emphasize the robustness and effectiveness of the proposed VME with dual-level contrastive learning as a superior approach for plant disease classification.

\subsection{Deployment Efficiency Discussion}

Contrastive learning frameworks generally involve data augmentation during training, which adds some computational overhead. However, this step is essential for learning robust and invariant representations.  In our framework, augmentations, such as colour jittering, rotation, and cropping, are computationally lightweight and can be efficiently applied using on-the-fly transformations. Importantly, once training is complete, the model operates without any data augmentation during inference. This allows VME to perform fast and low-latency predictions. The proposed framework is designed to be flexible: both augmentation intensity and model size can be adjusted to meet different hardware constraints. For example, in resource-constrained environments, lightweight Mamba variants (e.g., with fewer channels or reduced depth) can be integrated into the same contrastive learning pipeline and retrained efficiently without requiring architectural modifications. Furthermore, recent literature \cite{jayasimhan2024resprune,chen2023data}  has demonstrated that model compression techniques such as knowledge distillation and quantization can be effectively applied to Mamba-based architectures. These techniques present promising directions for deploying ConMamba on low-power platforms such NVIDIA Jetson or Raspberry. Exploring these directions represents a valuable area for future work.

\begin{table}[!t]
\centering
\caption{Performance comparison of Vision Mamba and traditional encoder using dual-level contrastive loss on the \textbf{Citrus} dataset. Evaluation metrics include Accuracy (\%), Recall (\%), Precision (\%), and F1-score (\%).}
    \vspace{-3pt}
\label{tab:ablation citrus}
\resizebox{0.6\linewidth}{!}{
\begin{tblr}{
  cells = {c},
  hline{1-2,10} = {-}{},
}
\textbf{Method}  & \textbf{Accuracy} & \textbf{Recall} & \textbf{Precision} & \textbf{F1-score} \\
AlexNet \cite{krizhevsky2017imagenet}         & 82.31             & 81.88           & 81.75              & 81.28             \\
MobileNetV3Large \cite{howard2019searching} & 80.85             & 76.13           & 76.57              & 81.44             \\
EfficientNetB0 \cite{howard2019searching}  & 83.16             & 82.13           & 81.46              & 81.74             \\
VGG16 \cite{simonyan2014very}           & 82.45             & 81.79           & 82.23              & 83.14             \\
ResNet50 \cite{he2016deep}        & 83.56             & 82.92           & 81.4               & 80.63             \\
ViT \cite{dosovitskiy2020image}             & 89.63             & 88.54           & 87.23              & 88.73             \\
Swin-T \cite{liu2021swin}             & 90.72             & 89.93           & 89.85               & 90.14             \\
\textbf{VM \cite{zhu2024vision}}              & \textbf{91.38}    & \textbf{90.51}  & \textbf{91.74}     & \textbf{91.66}    
\end{tblr}}
\caption*{\textbf{Note:} ViT = Vision Transformer; Swin-T = Swin Transformer; VM = Vision Mamba.}
\vspace{-12pt}
\end{table}
\subsection{Limitations}
Although ConMamba demonstrates strong performance on multiple benchmark datasets, its generalization capacity may still be affected by domain shift when applied to real-world scenarios involving unseen environmental conditions. Differences in lighting and background clutter characteristics may impact model reliability outside of the training distribution. Addressing this limitation in future work could involve exploring domain adaptation or domain generalization techniques to improve robustness across diverse deployment environments.

\vspace{-3pt}
\section{Conclusion}
\label{sec:Conclusion}
In this paper, we propose ConMamba, a novel self-supervised learning framework designed for plant disease detection (PDD), aiming to reduce dependence on large, annotated datasets. By leveraging unlabelled data, ConMamba learns rich and discriminative feature representations that improve classification performance across diverse and challenging scenarios. Specifically, ConMamba is based on the state-space model, which efficiently captures long-range dependent features. Long-range dependence is particularly important for PDD because disease symptoms can appear as subtle, scattered patterns across a leaf. By enhancing these long-range correlations, the model can integrate information from various regions of the leaves/plants. Thereby capturing the overall context and structure of the affected areas. Besides, we propose a dual-level contrastive learning strategy with a dynamic uncertainty-based loss weighting mechanism that aligns local and global features, enabling the model to effectively learn both fine-grained details and global structure. Extensive experiments on three benchmark datasets and comprehensive ablation studies validate the effectiveness and robustness of ConMamba. 
Future work will focus on refining the dynamic integration strategies, expanding the framework to additional datasets, and exploring its broader applications in precision agriculture and sustainable farming practices.

{
    \small
    \bibliographystyle{elsarticle-num}
    \bibliography{reference}
}

\end{document}